\documentclass[sigconf]{acmart}
\usepackage{graphicx}
\usepackage{amsmath}
\usepackage{caption}
\usepackage{enumerate}
\usepackage{multirow}
\usepackage{booktabs}
\usepackage{color}
\usepackage{subfig}
\usepackage{setspace}

\AtBeginDocument{%
  \providecommand\BibTeX{{%
    \normalfont B\kern-0.5em{\scshape i\kern-0.25em b}\kern-0.8em\TeX}}}

\copyrightyear{2022}
\acmYear{2022}
\setcopyright{acmcopyright}
\acmConference[MM '22] {Proceedings of the 30th ACM International Conference on Multimedia }{October 10--14, 2022}{Lisboa, Portugal.}
\acmBooktitle{Proceedings of the 30th ACM International Conference on Multimedia (MM '22), October 10--14, 2022, Lisboa, Portugal}
\acmPrice{15.00}
\acmISBN{978-1-4503-9203-7/22/10}

\copyrightyear{2022}
\acmDOI{10.1145/3503161.3547781}

\begin{document}

\title{Delving into the Frequency: Temporally Consistent Human Motion Transfer in the Fourier Space}


\author{Guang Yang}
% \authornote{This work was done when Guang Yang was an intern at Explore Academy of JD.com.}
\email{gyang@ict.ac.cn}
\authornotemark[1]
\authornotemark[3]
\affiliation{%
  \institution{Institute of Computing Technology, Chinese Academy of Sciences}
%   \institution{University of Chinese Academy of Sciences}
  \streetaddress{}
  \city{}
  \state{}
  \country{}
  \postcode{}
}

\author{Wu Liu}
\email{liuwu1@jd.com}
\authornotemark[2]
\affiliation{%
  \institution{JD Explore Academy}
  \streetaddress{}
  \city{}
  \state{}
  \country{}
  \postcode{}
}

\author{Xinchen Liu}
\email{liuxinchen1@jd.com}
\affiliation{%
  \institution{JD Explore Academy}
  \streetaddress{}
  \city{}
  \state{}
  \country{}
  \postcode{}
}

\author{Xiaoyan Gu}
\email{guxiaoyan@iie.ac.cn}
\affiliation{%
  \institution{Institute of Information Engineering, Chinese Academy of Sciences}
  \streetaddress{}
  \city{}
  \state{}
  \country{}
  \postcode{}
}

\author{Juan Cao}
\email{caojuan@ict.ac.cn}
\authornotemark[2]
\authornotemark[3]
\affiliation{%
  \institution{Institute of Computing Technology, Chinese Academy of Sciences}
%   \institution{University of Chinese Academy of Sciences}
  \streetaddress{}
  \city{}
  \state{}
  \country{}
  \postcode{}
}

\author{Jintao Li}
\email{jtli@ict.ac.cn}
% \authornotemark[3]
\affiliation{%
  \institution{Institute of Computing Technology, Chinese Academy of Sciences}
  \streetaddress{}
  \city{}
  \state{}
  \country{}
  \postcode{}
}

\thanks{*This work was done when Guang Yang was an intern at JD Explore Academy. \\ \dag Wu Liu and Juan Cao are corresponding authors. \\ \ddag The authors are at the Key Lab of Intelligent Information Processing of Chinese Academy of Sciences and the University of Chinese Academy of Sciences
% and the University of Chinese Academy of Sciences
.}

\renewcommand{\shortauthors}{Guang Yang et al.}

\begin{abstract}
Human motion transfer refers to synthesizing photo-realistic and temporally coherent videos that enable one person to imitate the motion of others.
However, current synthetic videos suffer from the temporal inconsistency in sequential frames that significantly degrades the video quality, yet is far from solved by existing methods in the pixel domain.
Recently, 
some works on DeepFake detection try to distinguish the natural and synthetic images in the frequency domain because of the frequency insufficiency of image synthesizing methods.
Nonetheless, there is no work to study the temporal inconsistency of synthetic videos from the aspects of the frequency-domain gap between natural and synthetic videos. 
Therefore, in this paper, we propose to delve into the frequency space for temporally consistent human motion transfer.
First of all, we make the first comprehensive analysis of natural and synthetic videos in the frequency domain to reveal the frequency gap in both the spatial dimension of individual frames and the temporal dimension of the video.
To close the frequency gap between the natural and synthetic videos, we propose a novel \textbf{Fre}quency-based human \textbf{MO}tion \textbf{TR}ansfer framework, named \textbf{FreMOTR}, which can effectively mitigate the spatial artifacts and the temporal inconsistency of the synthesized videos.
FreMOTR explores two novel frequency-based regularization modules: 
1) the Frequency-domain Appearance Regularization (FAR) to improve the appearance of the person in individual frames and 2) Temporal Frequency Regularization (TFR) to guarantee the temporal consistency between adjacent frames.
Finally, comprehensive experiments demonstrate that the FreMOTR not only yields superior performance in temporal consistency metrics but also improves the frame-level visual quality of synthetic videos.
In particular, the temporal consistency metrics are improved by nearly 30\% than the state-of-the-art model.
\end{abstract}






\maketitle

% \begin{small}
% \begin{spacing}{1}
% \textbf{ACM Reference Format:}
% % \blfootnote{*Wu Liu is the corresponding author.}

% \noindent Guang Yang, Wu Liu, Xinchen Liu, Xiaoyan Gu, Juan Cao, and Jintao Li. 2022. Delving into the Frequency: Temporally Consistent Human Motion Transfer in the Fourier Space. In \textit{Proceedings of the 30th ACM International Conference on Multimedia (MM'22), October 10–14, 2022, Lisbon, Portugal}. ACM, New York, NY, USA, 9 pages. https://doi.org/10.1145/3503161.3547781
% \end{spacing}
% \end{small}

\section{Introduction}
\label{sec:intro}
\begin{figure*}
    \includegraphics[width=0.95\textwidth]{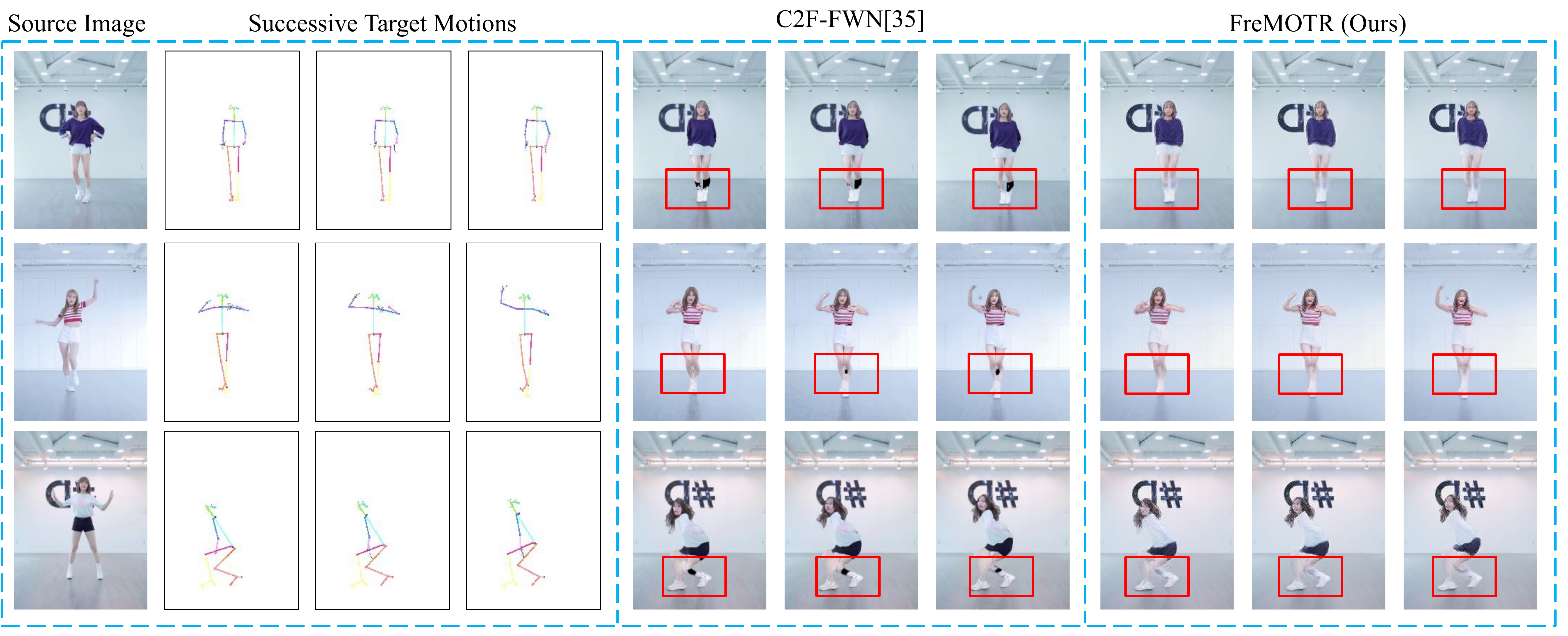}
    \caption{Comparison of successive frames synthesized by the basic human motion transfer model~(i.e., C2F-FWN~\cite{aaai/WeiXSH2021}) and our frequency-enhanced method FreMOTR, conditioned on source images and successive target motions. The FreMOTR reduces irrational components including artifacts and noise, which improves the temporal consistency between adjacent frames. Besides, as less noise exists in the frame, the FreMOTR also improves the quality of individual frames. (Zoom in for details.)}
    \label{fig:task_case}
\end{figure*}

Motion Transfer (HMT) aims to synthesize photo-realistic videos in which one person can imitate the motions of others~\cite{iccv/ChanGZE19}.
Specifically, given an image of a source person and a target video with the desired motion, the human motion transfer method can generate a video of the source person performing the target motion.
It is an emerging topic due to the potential applications like movie editing, virtual try-on, and online education~\cite{tpami/LiuPTLMG2021}.
Benefiting from the Generative Adversarial Networks~(GANs)~\cite{nips/GoodfellowPMXWOCB14, nips/Wang0TLCK19} with the pixel domain supervision, frame-level visual quality of some synthetic videos is significantly improved. 
However, as shown in Fig.~\ref{fig:task_case}, the results of current methods still suffer from artifacts and noise which lead to obvious temporal inconsistency and unsatisfied frame quality.
The temporal consistency of the synthetic videos is far from satisfactory yet neglected by the community, and thus this paper focuses on temporally consistent human motion transfer.

Most existing human motion transfer methods mainly focus on the appearance of persons in the pixel domain. 
In recent studies on the diagnosis of deep neural networks and GANs, frequency domain analysis has been proved to be an effective tool~\cite{icml/RahamanBADLHBC19}.
Some works have tried to utilize the frequency discrepancies between GAN-generated images and natural images for fake image detection~\cite{nips/DzanicSW20, icml/FrankESFKH20, nips/TancikSMFRSRBN20}.
They discovered that existing generative methods cannot effectively model the frequency-domain distribution of natural images, which causes the loss of information in specific frequencies in generated images.
Nonetheless, they simply adopt the frequency difference as loss, rather than make full use of the frequency domain characteristics. 
In this paper, we propose to regularize the appearance with frequency-aware networks to reduce the noise and improve the frame-level quality.

Moreover, recent human motion transfer methods mainly focus on frame-level quality with precise human structure representations such as 2D human pose~\cite{cvpr/BalakrishnanZDD18, cvpr/Tulyakov0YK18}, human parsing~\cite{nips/DongLGLZY18, tog/LiuXZKBHWT19, mm/LiuZLSM19}, 3D human pose~\cite{nips/Wang0ZYTKC18}, and other 3D informationl~\cite{aaai/liu2018t, liu2021recent, sun2022putting, Zheng_2022_CVPR, iccv/LiuPML0G19, tpami/LiuPTLMG2021}.
Although some methods try to use a multi-frame smoothing loss~\cite{nips/Wang0TLCK19} or optical flow~\cite{aaai/WeiXSH2021} in the pixel domain to improve the temporal consistency, the temporal artifacts still cannot be effectively eliminated in the pixel domain.
These temporal artifacts can be easily perceived by the human visual system and significantly degrade the quality of videos~\cite{tip/SeshadrinathanB10}.
However, there is seldom work to study whether the temporal inconsistency of synthetic videos arises from the frequency-domain gap between natural videos and synthetic videos, let alone how to eliminate them in the frequency domain. 

In this paper, we delve into the Fourier space for temporally consistent human motion transfer from two perspectives as shown in Fig.~\ref{fig:workflow}.
First, we explore a Frequency-domain Appearance Regularization~(FAR) method that models the frame-level characteristics in both pixel and frequency domains to reduce noise in individual frames.
Second, we discover and close the temporal frequency domain gap between natural videos and synthetic videos by a Temporal Frequency Regularization~(TFR) module.
By reducing noise in individual frames and improving the long-term coherency, both the frame-level quality and temporal consistency of generated motion videos are improved. 

First, we synthesize coarse frames with HMT backbones and regularize the appearance for a finer appearance.
The FAR works on the single coarse frames.
To improve the frequency insufficiency of current generative method, Fast Fourier Convolution is adopted instead of the traditional convolution block to focus on characteristics in both pixel and frequency domain.
By regularizing the original image in the frequency domain and then reconstructing the image, FAR reduces noise and artifacts in the frame.

The TFR is designed with the observation that the temporal changes in amplitude and phase spectra of synthetic videos are significantly different from the ones of natural videos, which is shown in Fig.~\ref{fig:tfc}. This phenomenon demonstrates that existing methods cannot reproduce the frequency-domain distribution as natural videos using only pixel-domain supervision.
To this end, we design a Weighted Temporal Frequency Regularization (WTFR) loss for human motion transfer methods, which applies regularization to make the temporal changes between adjacent frames of synthetic videos approximate to natural videos in the frequency domain.
In particular, the WTFR loss is applied to the amplitude and phase spectrum separately, because the temporal variations of amplitude and phase demonstrate different patterns for natural videos and synthetic videos~\cite{tip/SeshadrinathanB10}. 
For example, the phase spectrum of natural videos has more changes in the high-frequency components compared with the amplitude spectrum, as the spectra changes of natural videos shown in Fig.~\ref{fig:tfc}.
More specifically, to alleviate the fluctuation of high-frequency components, we add a frequency-aware weighting term in the loss function.
The WFTR loss can be integrated into existing human motion transfer methods without any modification to model structures.
With the WTFT loss, these methods can learn to close the temporal frequency discrepancy between synthetic and natural videos.

Both quantitative and qualitative evaluations show that the temporal consistency of synthetic videos is significantly improved, as shown in Fig.~\ref{fig:task_case}. 
Our main contributions are summarized as follows:
\vspace{-5mm}
\begin{enumerate}[1)]
    \item To the best of our knowledge, we make the first attempt to reveal the temporal gap between natural videos and synthetic videos in the frequency domain.
    \item More importantly, we build the Temporal Frequency Prior into deep networks through a well-designed WTFR loss, which can effectively guide the GANs to model the temporal frequency changes of natural videos.
    \item To reduce the noise and artifacts in frames, the FAR is proposed to regularize the generation process in both pixel and frequency domains.
\end{enumerate}
\vspace{-3mm}
Experiments show that our method can effectively improve the temporal consistency of the generated videos. 
In particular, the Temporal Consistency Metrics~\cite{mm/YaoCC17} are improved by nearly 30\% than the state-of-the-art models.
The frame-level metrics are also improved a bit, while the qualitative analyses demonstrate that the improvement is significant for visual perception.

\section{Related Work}

\subsection{Human Motion Transfer}
Human motion transfer has been studied for over two decades.
Early methods are focused on retrieval-based pipelines to retarget the subject in a video to the desired motion with optical flow~\cite{iccv/EfrosBMM03} or 3D human skeleton~\cite{tog/XuLSTBDSKT11}.
Recent methods usually adopt deep neural networks, especially GANs~\cite{nips/GoodfellowPMXWOCB14}, and exploit 2D human pose~\cite{cvpr/Tulyakov0YK18, cvpr/BalakrishnanZDD18, cvpr/SiarohinLT0S19, iccv/ChanGZE19, nips/Wang0TLCK19}, human parsing~\cite{nips/DongLGLZY18, tog/LiuXZKBHWT19, mm/LiuZLSM19, tmm/WeiSH2020}, 3D human pose~\cite{nips/Wang0ZYTKC18}, and other 3D information~\cite{aaai/liu2018t, liu2021recent, sun2022putting, Zheng_2022_CVPR} for human motion transfer~\cite{iccv/LiuPML0G19, tpami/LiuPTLMG2021}.

However, most of these methods only model the appearance mapping between individual source and target frames in the pixel domain, while the frequency domain and temporal consistency are usually neglected.
Although some methods like vid2vid~\cite{nips/Wang0ZYTKC18, nips/Wang0TLCK19} and EDN~\cite{iccv/ChanGZE19} take multiple frames as the input to improve temporal coherence in the pixel domain for temporal smoothing, the synthesized motion videos are still with obvious temporal artifacts.
Most recently, \citet{aaai/WeiXSH2021} proposed C2F-FWN with a Flow Temporal Consistency Loss to explicitly constrain temporal coherence by optical flow, but optical flow estimation could bring high extra computation cost and noise.
Therefore, this paper focuses on exploring a more effective method to explicitly improve the temporal consistency for human motion transfer.

\vspace{-3mm}
\subsection{Frequency Domain Analysis for GANs}
In recent research, frequency domain analysis for deep neural networks, especially GANs, has been proved an effective tool for network diagnose~\cite{icml/RahamanBADLHBC19, cvpr/DurallKK20}, GAN generated image detection~\cite{wifs/0022KC19, icml/FrankESFKH20, nips/DzanicSW20}, and improving the capability of GANs for image generation~\cite{nips/TancikSMFRSRBN20, aaai/ChenLJLL2021}.
% \cite{corr/abs-1911-00686} showed that synthetic images are different from real images in high-frequency coefficients.
For example, ~\citet{icml/RahamanBADLHBC19} used Fourier analysis to highlight the bias of deep networks towards low-frequency functions.
~\citet{cvpr/DurallKK20} discovered that the up-sampling or up-convolnution operations made GANs fail to reproduce the spectral distributions of natural images.
\citet{nips/DzanicSW20} explored the Fourier spectrum discrepancies of high-frequency components between natural and generated images for fake image detection.
\citet{aaai/ChenLJLL2021} observed missing of high-frequency information in discriminators of GANs.
% They also proposed spectral regularization terms in the losses of GANs which could improve the quality of generated images.
Nevertheless, there is no prior work to study whether the temporal inconsistency of synthetic videos is caused by the frequency discrepancies between natural videos and synthetic videos.

\vspace{-3mm}
\subsection{Video Temporal Consistency}
Temporal consistency is an essential aspect for video quality assessment~\cite{tip/SeshadrinathanB10}, video processing~\cite{mm/YaoCC17, cvpr/LiuLZCGYM19}, video frame prediction~\cite{nips/BhattacharjeeD17}, and blind video temporal consistency~\cite{tog/BonneelTSSPP15, eccv/LaiHWSYY18, nips/LeiXC20}.
For example, \citet{tip/SeshadrinathanB10} designed the motion-based video integrity evaluation index, which adopted Gabor filters in the frequency domain for evaluating dynamic video fidelity.
\citet{mm/YaoCC17} proposed an occlusion-aware temporal warping algorithm to achieve temporal consistency for video processing.
\citet{nips/BhattacharjeeD17} proposed a temporal coherency criterion for video frame prediction using GANs.
\citet{nips/LeiXC20} proposed to learn a Deep Video Prior from natural and processed videos and achieved superior temporal consistency on seven video processing or generation tasks.
This paper presents a novel temporal consistency regularization loss using Temporal Frequency Prior obtained by Fourier analysis.
It can be used for existing human motion transfer methods without any modification on models and exhibit state-of-the-art performance.

\section{FreMOTR}

\subsection{Fourier Transform of Image}
\label{subsec:DFT}
The Fourier Transform convert the image in pixel domain into the frequency domain.
In our study, we adopt the 2D-Discrete Fourier Transform (2D-DFT) to obtain the frequency spectrum of a frame.
Specifically, given a frame represented by a discrete 2D signal $f(x, y)\in \mathbb{R}^{M\times N}$, its frequency spectrum value at coordinate $(u,v)$ is obtained by
\begin{equation}
\label{eq:DFT}
F(u,v)=\sum_{x=0}^{M-1} \sum_{y=0}^{N-1} f(x,y) \cdot e^{-i 2 \pi\left(\frac{u x}{M}+\frac{v y}{N}\right)}.
\end{equation}
The frequency spectrum can be denoted as $F(u,v)=R(u,v)+I(u,v)i$ where $R(u,v)$ is the real part and $I(u,v)$ is the imaginary part.
The amplitude that reflects the intensity is obtained by 
$|F(u,v)|=\sqrt{R(u,v)^2+I(u,v)^2}$, while the phase that reflects the position is denoted by $\angle F(u,v)=\arctan(I(u,v)/R(u,v))$.
Hence the frequency spectrum can be divided into the amplitude spectrum and the phase spectrum.
For visualization in the following figures, both spectra are shifted so that the centers denote low-frequency components while the corners denote the high-frequency components, and the amplitude spectrum is further converted into logarithmic units. 

\begin{figure*}[!htpb]
    \centering
    \includegraphics[width=0.98\textwidth]{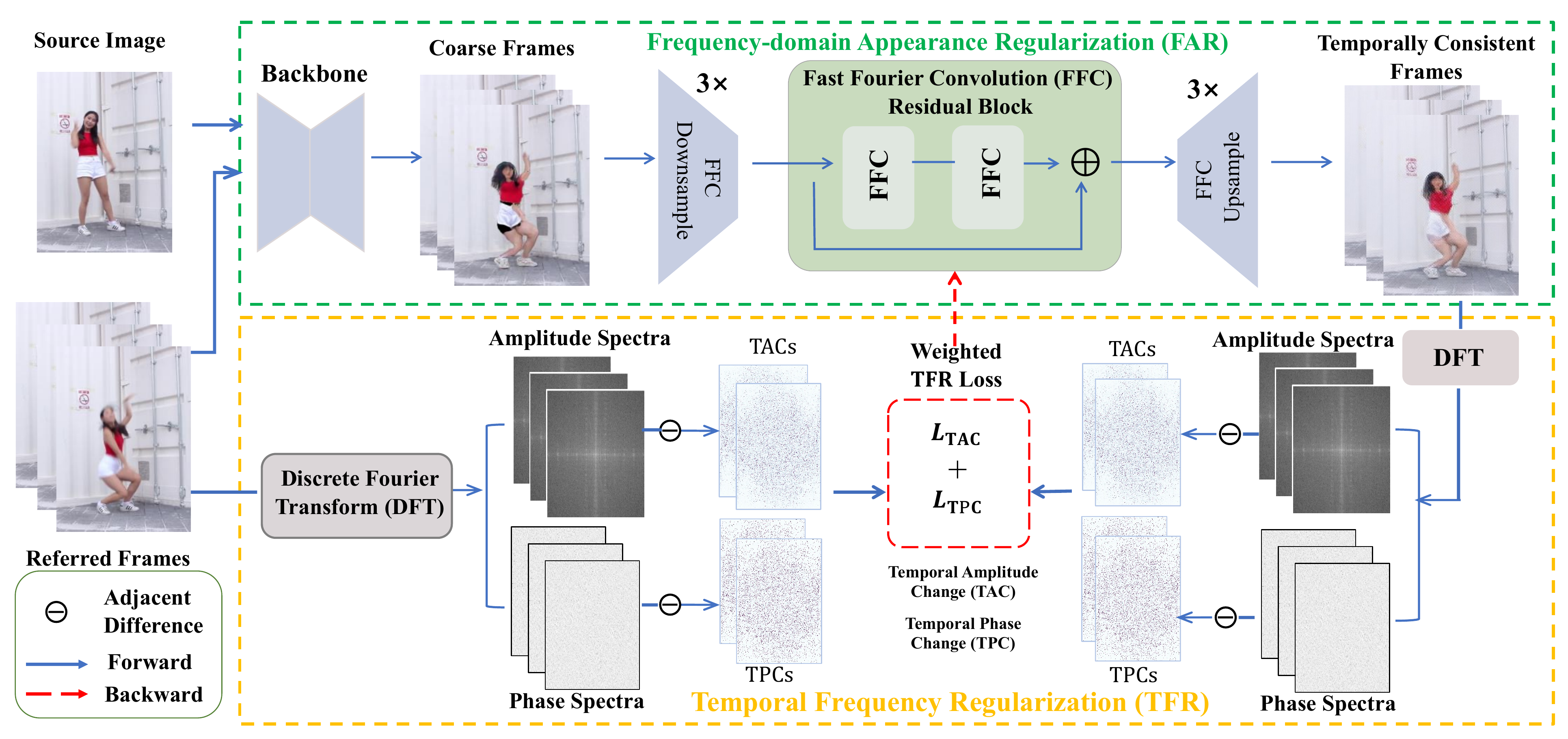}
    \caption{Overview of the proposed FreMOTR for temporally consistent human motion transfer, which contains two main parts, i.e. Frequency-domain Appearance Regularization~(FAR) and Temporal Frequency Regularization~(TFR). First, FAR regularize the coarse frames synthesized by backbone in both pixel and frequency domain with the Fast Fourier Convolution~(FFC). Second, the TFR transforms the frames into the Fourier Space with Discrete Fourier Transform~(DFT), and calculate the Temporal Amplitude Change~(TAC) and the Temporal Phase Change~(TPC) conditioned on the adjacent difference of the amplitude and phase spectra, respectively. The Weighted Temporal Frequency Fegularization~(WTFR) loss combines the weighted differences of TAC and TPC between referred and synthetic frames~(i.e., $L_{TAC}$ and $L_{TPC}$)). Finally, the WTFR loss guides the model to learn to approximate the frequency distribution of natural videos. }
    \label{fig:workflow} \vspace{-3mm}
\end{figure*} 

\subsection{Frequency-domain Appearance Regularization}

The incoherent appearance like noise and artifacts can be easily perceived by the human visual system and significantly degrade the quality of videos~\cite{tip/SeshadrinathanB10}. However, these signals cannot be easily described in the pixel  domain. To catch and reduce these signals, we process the image in the frequency domain with Fast Fourier Convolution~(FFC)~\cite{nips/chi2020}. In addition, FFC can catch the global information without the limitation of receptive field, which will benefit the coherency of the appearance.

\subsubsection{Fast Fourier Convolution}
\begin{figure}[t]
    \centering
    \includegraphics[width=1.0\columnwidth]{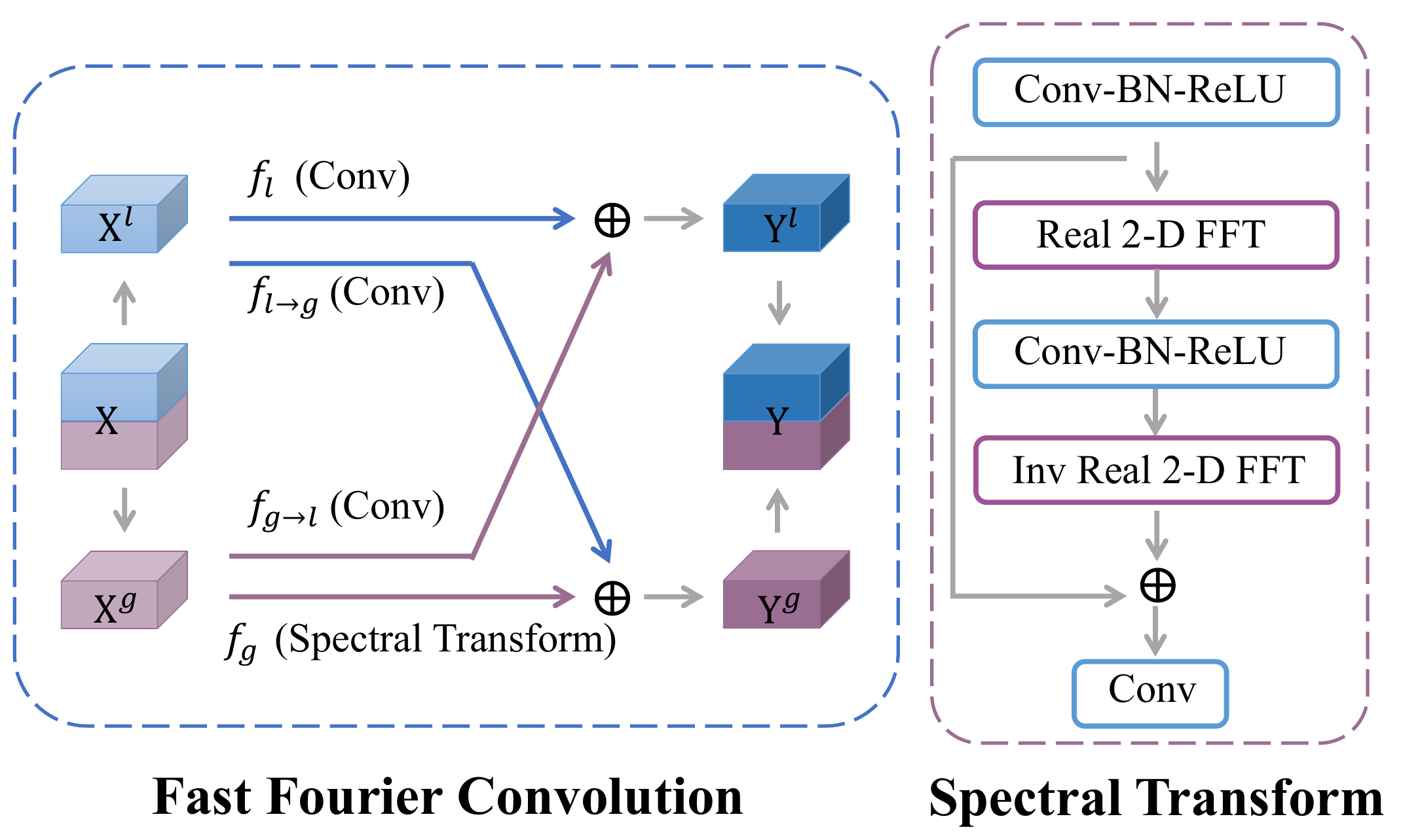}
    \caption{Workflow of the Fast Fourier Convolution~(FFC)}
    \label{fig:FFC} \vspace{-5mm}
\end{figure}

% \ \\
The workflow of a FFC block is presented in Fig.~\ref{fig:FFC}, which contains two main streams.The input $X$ is first divided into two parts. The top stream adopts traditional convolution to capture local information~(i.e., $Y^l$). The bottom stream processes the feature in the frequency domain~(i.e., Spectral Transform) to capture frequency information and global information~(i.e., $Y^g$). Two streams complement each other by exchanging local and global information as follows: 
\begin{equation}
\begin{aligned}
\mathbf{Y}^{l} &=\mathbf{Y}^{l \rightarrow l}+\mathbf{Y}^{g \rightarrow l}=f_{l}(\mathbf{X}^{l})+f_{g \rightarrow l}\left(\mathbf{X}^{g}\right) \\
\mathbf{Y}^{g} &=\mathbf{Y}^{g \rightarrow g}+\mathbf{Y}^{l \rightarrow g}=f_{g}\left(\mathbf{X}^{g}\right)+f_{l \rightarrow g}(\mathbf{X}^{l})
\end{aligned}
\end{equation}
In the Spectral Transform~($f_g$), as shown in the right figure of Fig.~\ref{fig:FFC}, the feature is processed sequentially with Real 2-D FFT, frequency domain convolution, and Inverse Real 2-D FFT.

\subsubsection{Regularize Appearance with FFC}
%\ \\
As shown in the top of Fig.~\ref{fig:workflow}, we regularize the appearance in the frequency domain with FFC blocks. The Frequency-domain Appearance Regularization~(FAR) is trained to reduce the noise and artifacts in the coarse frame to make the appearance consistent for adjacent frames.
First, the coarse frames are downsampled with 3 FFC blocks. Then the frame features are processed with several FFC residual blocks in both pixel and frequency domain, aiming to reduce the artifacts and noise which are abnormal or discontinuous. At last, the final features are upsampled with 3 FFC blocks to reconstruct the frame.

\subsection{Temporal Frequency Regularization}

For natural videos, short segments of video without any scene changes contain local image patches undergoing translation, which will lead to major phase changes and minor amplitude changes in the frequency domain~\cite{tip/SeshadrinathanB10}. Inspired by the flaw of GANs in the frequency domain, we discover the temporal frequency gap and build a targeted regularization to close the gap in this part.

\subsubsection{Temporal Frequency Changes of Adjacent Frames}
%\ \\
To discover the gap between natural and synthetic videos, we analyze the temporal changes of adjacent frames in amplitude and phase separately, which is shown in the bottom of Fig.~\ref{fig:workflow}. 
We define the Temporal Frequency Change (TFC), i.e., the Temporal Amplitude Change~(TAC) of $|F(u,v)|$ and Temporal Phase Change~(TPC) of $\angle F(u,v)$ between adjacent frames $f_t$ and $f_{t-1}$ in a video, as
\begin{equation}
\label{eq:tfc}
\begin{aligned}
{TAC_t}(u,v)&=||F_t(u,v)|-|F_{t-1}(u,v)||,\\
{TPC_t}(u,v)&=|\angle F_t(u,v)-\angle F_{t-1}(u,v)|.
\end{aligned}
\end{equation}

% \begin{figure}[t]
%     \centering
%     \includegraphics[width=1.0\columnwidth]{Figures/TFA.pdf}
%     \caption{Workflow of Temporal Frequency Analyzing}
%     \label{fig:TFA}
% \end{figure}

\begin{figure}[t]
    \centering
    \includegraphics[width=1.0\columnwidth]{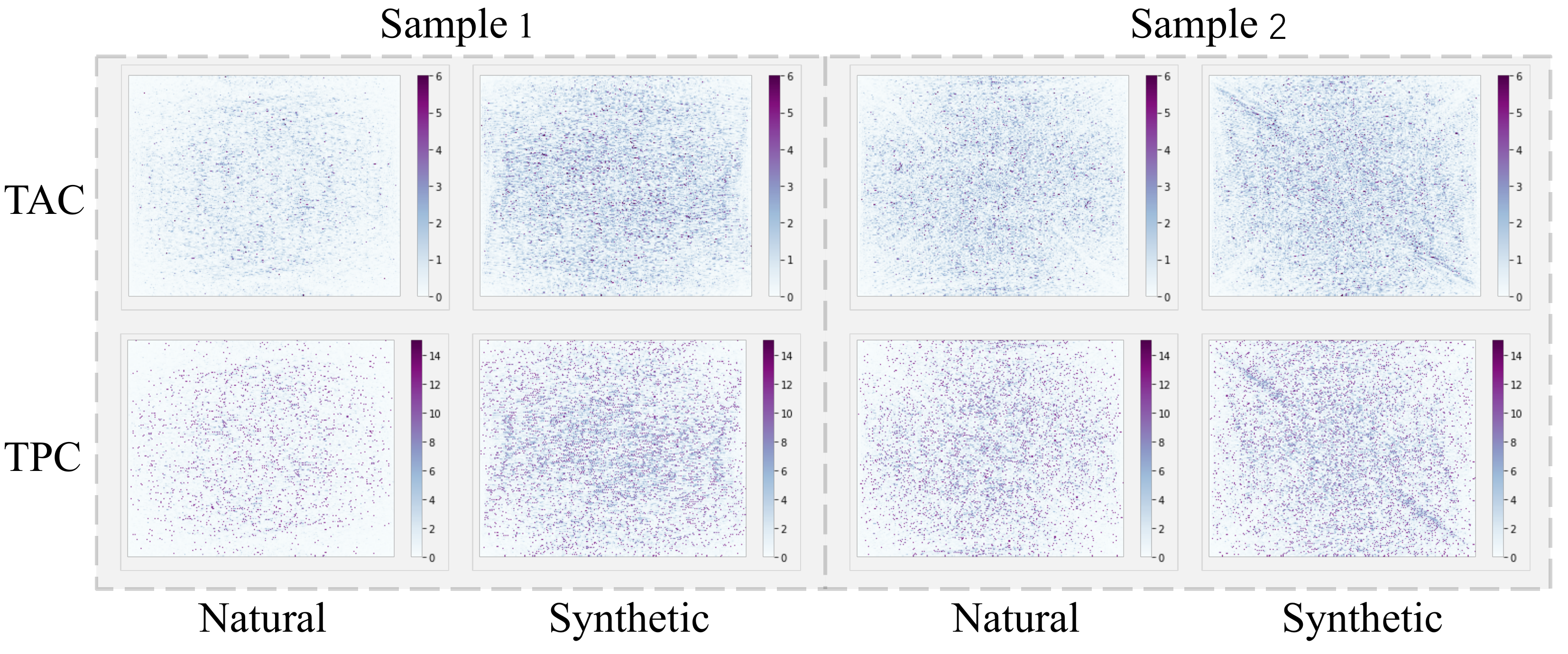}
    \caption{Example heatmaps of Temporal Frequency Change~(i.e., TAC and TPC) from natural videos and corresponding synthetic videos. The TAC and TPC of synthetic videos are often with larger values, especially in the high-frequency components.}
    \label{fig:tfc} \vspace{-5mm}
\end{figure}

We analyzed the synthetic frames with low temporal consistency and drawn several observations, and several example heatmaps of TFC are shown in Fig.~\ref{fig:tfc}.
First, for adjacent synthetic frames with low temporal consistency, TFCs scatter over various frequencies with larger values, while the ones of natural videos are mainly in the low-frequency components with smaller values.
Second, the phase spectrum of natural videos contains more changes in the high-frequency components compared with the amplitude spectrum. This observation is consist with the conclusions in~\cite{tip/SeshadrinathanB10}. The main content~(i.e., intensity of the pixels) of adjacent frames does not change too much, which is reflected in the amplitude strum. The position of the pixels change but not very intensely, which is reflected in the phase spectrum.

\subsubsection{Weighted Temporal frequency Regularization}
\label{subsubsec:WTFR}
%\ \\
Based on the above observation, we present how to utilize the findings for temporally consistent human motion transfer.
The overview of the proposed framework is shown in bottom of Fig.~\ref{fig:workflow}.
For any GAN-based human motion transfer model, we first calculate the temporal frequency changes (TFCs) between adjacent frames of referred videos and generated videos.
Then a Weighted Temporal Frequency Regularization (WTFR) loss is obtained from the TFCs to guide the training of the model. 
By this means, the model can make the temporal patterns of generated videos approximate to natural videos in the frequency domain.

Given a source image $I^S$ and a frame sequence $\{f^R_t\}^T$ with the referenced motion, the model aims to synthesize a frame sequence $\{f^S_t\}^T$ conditioned on the appearance in $I^S$ and the motion in $\{f^R_t\}^T$, where $T$ is the length of the referenced sequence.
Motions in adjacent frames exhibit temporal frequency changes which reflect different magnitudes in amplitude and phase spectra.
To make the model learn the TFC of natural videos, we define the WTFR loss, which regularizes TAC and TPC of adjacent frames separately.
For TAC, the loss function $L_{TAC}$ is formulated by
\begin{equation}
\label{eq:L_TAC}
L_{TAC}(f^R_t\!,\!f^S_t)=\frac{1}{M\!N}\!\sum_{u=0}^{M-1}\! \sum_{v=0}^{N-1}\!w(u,v)*|TAC^R_t\!-\!TAC^S_t|,
\end{equation}
and the loss function $L_{TPC}$ for TPC is formulated by
\begin{equation}
\label{eq:L_TPC}
L_{TPC}(f^R_t\!,\!f^S_t)=\frac{1}{M\!N}\!\sum_{u=0}^{M-1}\! \sum_{v=0}^{N-1}\!w(u,v)*|TPC^R_t\!-\!TPC^S_t|,
\end{equation}
where $w(u,v)=[(\frac{M}{2})^2+\frac{N}{2})^2]^\delta-[(u-\frac{M}{2})^2+(v-\frac{N}{2})^2]^\delta+1$ is a weighting term to penalize high-frequency components based on the observation that TFCs in natural videos are relatively small in high-frequency components as shown in Fig.~\ref{fig:tfc}). $\delta$ is a hyper-parameter to scale the magnitude.

Finally, the WTFR loss is denoted as
\begin{equation}
\label{eq:wtfr}
L_{WTFR}(f^R_t\!,\!f^S_t)\!=\!\alpha*L_{TAC}(f^R_t\!,\!f^S_t)\!+\!\beta*L_{TPC}(f^R_t\!,\!f^S_t),
\end{equation}
where $\alpha$ and $\beta$ are hyper-parameters. With the WTFR loss, the model can learn to approximate the frequency distribution of natural videos by minimizing the $L_1$ distance of TFCs between the referenced video and the synthetic video.

\section{Experiments}
In this section, we conduct experiments to answer the following evaluation questions:

% \textbf{EQ1: }Can FreMOTR improve the metrics of temporal consistency as well as the frame-level image quality? (in Sec.~\ref{subsec:quantitative})

% \textbf{EQ2: }Does FreMOTR indeed reduce the inconsistent and unnatural components from the visual perception aspect? (in Sec.~\ref{subsubsec:eq2})

% \textbf{EQ3: }Can FreMOTR reduce the discovered temporal frequency gap? (in Sec.~\ref{subsubsec:eq3})

% \textbf{EQ4: }How each components improve the temporal consistency? And for temporal frequency regularization, are amplitude and phase spectrum both helpful? (in Sec.~\ref{subsec:ablation})

\begin{itemize}
    \item[\textbf{EQ1}] Can FreMOTR improve the metrics of temporal consistency as well as the frame-level image quality? (in Sec.~\ref{subsec:quantitative})
    \item[\textbf{EQ2}] Does FreMOTR indeed reduce the inconsistent and unnatural components from the visual perception aspect? (in Sec.~\ref{subsubsec:eq2})
    \item[\textbf{EQ3}] Can FreMOTR reduce the discovered temporal frequency gap? (in Sec.~\ref{subsubsec:eq3})
    \item[\textbf{EQ4}] How each components improve the temporal consistency? And for temporal frequency regularization, are amplitude and phase spectrum both helpful? (in Sec.~\ref{subsec:ablation})
\end{itemize}
\vspace{-3mm}

\subsection{Experimental Setups}
\label{subsec:setup}

\subsubsection{Dataset}
%\ \\
% We adopted two datasets~(i.e., iPER and SoloDance) to validate the proposed framework. The results of iPER are provided in the appendix.
% \indent \textbf{iPER}~\cite{cvpr/LiHL19} The released dataset contains 206 videos and 241,564 frames captured by the authors. The ration of training and testing set is 8:2. There are 30 subjects with different shapes, heights, genders, and clothes. Each subject is asked to perform an A-pose video and a video with random actions like Tai Chi and arm exercises.
% The LWGAN is trained on iPER using the settings in~\cite{iccv/LiuPML0G19, tpami/LiuPTLMG2021}. 
\textbf{SoloDance}~\cite{aaai/WeiXSH2021} 
We adopted C2F-FWN as our backbone, and thus we adopted the SoloDance dataset released by C2F-FWN for comparison.
The SoloDance dataset contains 179 solo dance videos with 53,700 frames collected from the Internet.
There are 143 human subjects with various clothes and performing complex solo dances in various backgrounds. 
This dataset is split into 153 and 26 videos for training and testing, respectively. 
% We train the C2F-FWN model using the settings in~\cite{aaai/WeiXSH2021}. 

\subsubsection{Methods for Comparison}
%\ \\
To validate the effectiveness of the FreMOTR, we compare the performance with the following HMT methods: FSV2V~\cite{nips/Wang0TLCK19} that takes multi frames as input to improve temporal coherence, SGWGAN~\cite{nips/DongLGLZY18} and LWGAN~\cite{iccv/LiuPML0G19} that warp the features, ClothFlow~\cite{iccv/han2019} that warp the image, and C2F-FWN~\cite{aaai/WeiXSH2021} that constrains temporal coherence by optical flow. 
\begin{figure*}[t]
    \centering
    \includegraphics[width=1.0\textwidth]{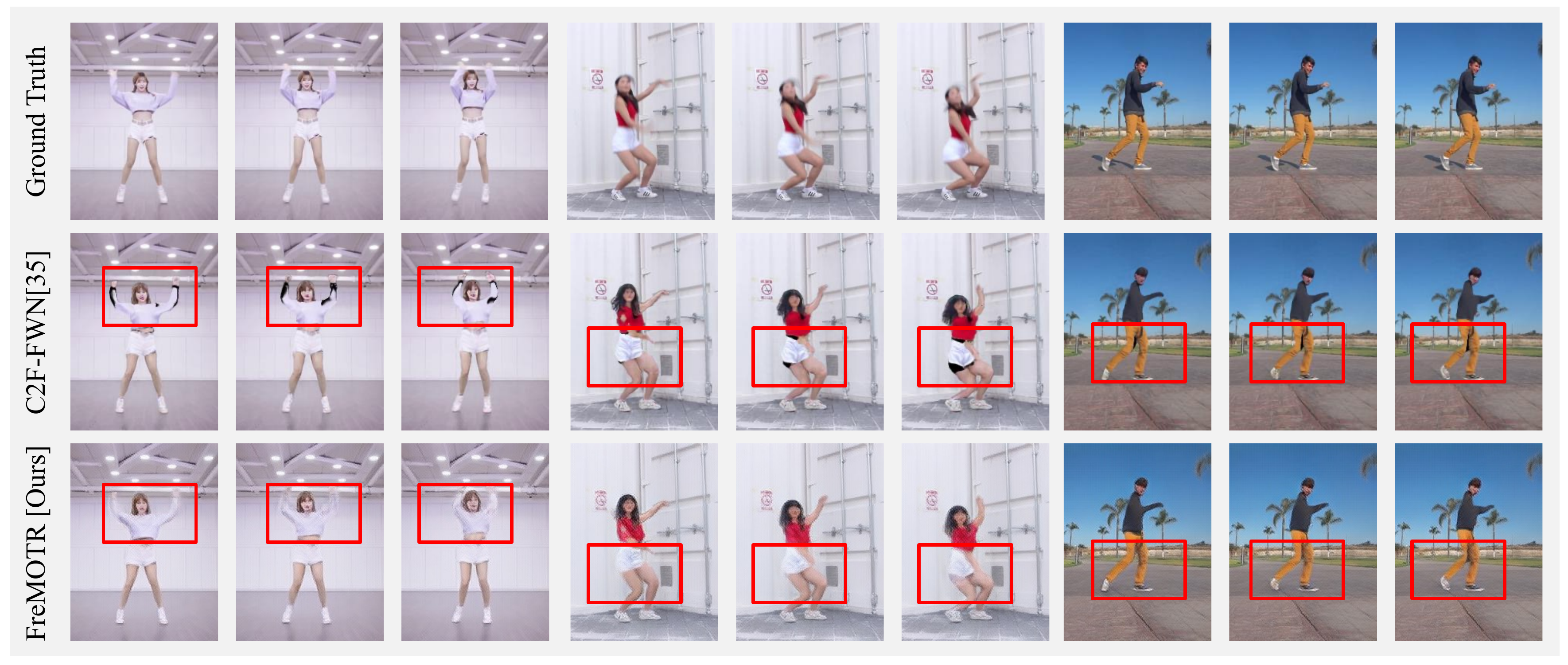}
    \caption{Visual performance of various successive synthetic frames. From top to bottom are ground truth, results from C2F-FWN~\cite{aaai/WeiXSH2021}, and results from the proposed FreMOTR. More results are in the appendix.}
    \label{fig:case_study} \vspace{-3mm}
\end{figure*}
The C2F-FWN is the backbone in our experiments, but the authors did not release pretrained weights.
Therefore, we trained the C2F-FWN based on the released code and settings, and report the reproduced performance. Performances of other methods are from ~\cite{aaai/WeiXSH2021}.

\subsubsection{Implementation Details}
%\ \\
We adopt the C2F-FWN that achieves the SOTA temporal consistency as the backbone for the SoloDance dataset. We adopt a three-stage training strategy. During the first stage, we train the backbone using the settings and hyper-parameters in the released codes of C2F-FWN until converge, then the parameters are fixed. During the second stage, the FAR is combined and trained with general image generation loss like adversarial loss~\cite{nips/GoodfellowPMXWOCB14} and perceptual loss~\cite{eccv/Johnson16} until converge. Finally, the WTFR is added to the general losses to finetune the FAR.

The experiments are conducted on four NVIDIA Tesla P40 GPUs with 24GB memory based on PyTorch~\cite{nips/paszke2019}. 
Following C2F-FWN, the input frames are resized and cropped to 192$\times$256.
The Adam optimizer~\cite{iclr/KingmaB14} is used for optimization with a batch size of 8.
The hyper-parameters $\alpha$, $\beta$, and $\delta$ for the WTFR loss in Eq.~\ref{eq:wtfr} are set to $0.5$, $1$, and $0.05$, respectively, to make the WTFR has the same order of magnitude of other losses.
The FAR has 3 downsampling blocks, 3 upsampling blocks, and 18 residual blocks. The initialization of FAR is based on the pretrained FFC provided by~\citet{wacv/Suvorov2022}.

We also adopt only WTFR for ablation. After the first stage that the backbone is well trained, the WTFR is used to finetune the backbone until converge.

\subsubsection{Evaluation Metrics}
\label{subsec:metrics}
%\ \\
To measure the frame-level image quality, we adopt SSIM~\cite{tip/wang04}, PSNR, and LPIPS~\cite{cvpr/zhang2018} following previous works~\cite{aaai/WeiXSH2021}. 
Specifically, SSIM, PSNR, and LPIPS compare the pairwise similarity of synthetic frames and ground-truth frames.
Higher value is better for SSIM and PSNR, while lower value is better for LPIPS.
The frame-level metrics are averaged over all frames of all synthetic videos.

To evaluate the temporal consistency, we adopt Temporal Consistency Metric~(TCM)~\cite{mm/YaoCC17}.
TCM is calculated from the warping error between frames. 
We first obtain the optical flow $o_t$ between adjacent referenced frames $f^R_t$ and $f^R_{t-1}$ using Flownet~2.0~\cite{cvpr/ilg2017}.
The optical flow is used to obtain two adjacent frames by warping $f^S_{t-1}$ and $f^R_{t-1}$. 
Finally, TCM is obtained by the ratio of warping errors as  
\begin{equation}
\label{eq:TCM}
{TCM}_{t}=\exp \left(-\left|\frac{\left\|f^R_{t}-\operatorname{warp}\left(f^R_{t-1}, o_t\right)\right\|^{2}}{\left\|f^S_{t}-\operatorname{warp}\left(f^S_{t-1}, o_t\right)\right\|^{2}}-1\right|\right),
\end{equation}
where $\operatorname{warp}(\cdot, \cdot)$ is the warping operation.
The temporal metrics are averaged over all adjacent frames of all synthetic videos.

\subsection{Quantitative Analysis~(EQ1)}
\label{subsec:quantitative}
\vspace{-1mm}
The quantitative metrics of the results are presented in Table~\ref{tab:quantitative}. In general, compared with all other methods
% including the backbone~(i.e., C2F-FWN)
, the proposed FreMOTR improves both temporal consistency and frame-level image quality.
Specifically, the FreMOTR significantly improves the TCM by about 30\% compared with the state of the art. In addition, FreMOTR improves the frame quality metric~(i.e., LPIPS and PSNR). 
The quantitative analysis reveals that our frequency-domain method can effectively improve existing methods that are only optimized by pixel-domain supervision.
Although SSIM of FreMOTR is not the best, such frame-quality metrics only evaluate specific aspects and are for reference. The results of FreMOTR have fewer artifacts and unnatural patches in the synthetic image, which are demonstrated in the qualitative analysis.

\begin{table}[t]
\centering
\caption{Quantitative evaluation of FreMOTR on the SoloDance dataset. TCM is a temporal consistency metric, while LPIPS, PSNR, and SSIM are frame-level image quality metrics. $\uparrow$ means higher is better, while $\downarrow$ means lower is better. The best and second-best results in each column are \textbf{boldfaced} and {underlined}, respectively.}  \vspace{-3mm}
\label{tab:quantitative}
\begin{tabular}{@{} c l l l l l l @{}}
\toprule
\multirow{2}{*}{HMT method} & & \multicolumn{1}{c}{{Temporal}} && \multicolumn{3}{c}{Frame Quality}\\
\cmidrule{3-3} \cmidrule{5-7}
&& TCM $\uparrow$ && LPIPS $\downarrow$ &PSNR $\uparrow$ & SSIM $\uparrow$ \\
\midrule
FSV2V~\cite{nips/Wang0TLCK19}  & & 0.106 && 0.132 & 20.84 & 0.721 \\
SGWGAN~\cite{nips/DongLGLZY18} & & 0.166 && 0.124 & 20.54 & 0.763 \\
ClothFlow~\cite{iccv/han2019} & & 0.322 && 0.072 & 22.06 & \textbf{0.843} \\
LWGAN~\cite{iccv/LiuPML0G19} & & 0.176 && 0.106 & 20.87 & 0.786 \\
C2F-FWN~\cite{aaai/WeiXSH2021} & & \underline{0.523} && \underline{0.060} & \underline{22.17} & \underline{0.821} \\
FreMOTR & & \textbf{0.678} && \textbf{0.059} & \textbf{22.25} & 0.780 \\
\bottomrule
\end{tabular} \vspace{-5mm}
\end{table}

\begin{figure*}[t]
    \centering
    \includegraphics[width=1.0\textwidth]{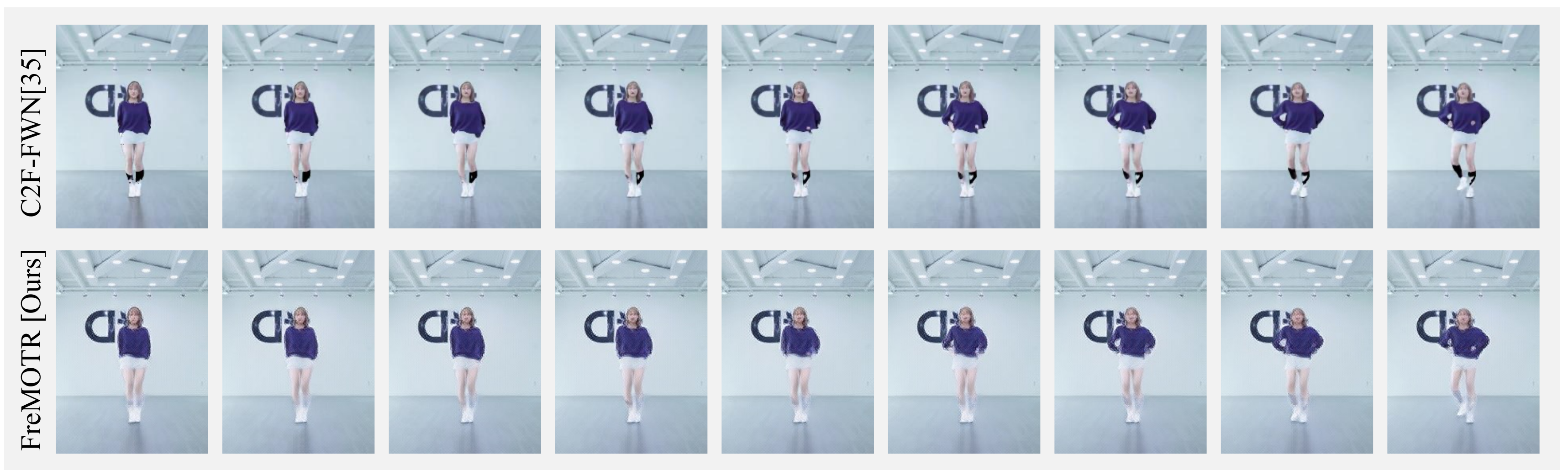}
    \caption{Visual quality of long-time successive frames. The top results are from C2F-FWN~\cite{aaai/WeiXSH2021} with obvious inconsistent artifacts. The bottom with long-time consistency are from the proposed FreMOTR. More results are in the appendix.}
    \vspace{-3mm}
    \label{fig:long_time} 
\end{figure*}

\subsection{Qualitative Analysis~(EQ2\&3)}
\label{subsec:qualitative}
\vspace{-1mm}

\subsubsection{Improvement of the Visual Performance~(EQ2)}
\label{subsubsec:eq2}
%\ \\
We present some of the synthesized frames in Fig.~\ref{fig:case_study}, from top to bottom are ground truth, results from C2F-FWN, and results from our FreMOTR. The results of C2F-FWN often contain artifacts, and the consistency of the appearance cannot be preserved even in three successive frames. 

The visual results demonstrate that with the FFC that mitigates the incoherent appearance and the WTFR that closes the temporal frequency gap, FreMOTR significantly reduces the inconsistency between successive frames. Therefore, the FreMOTR can achieve a long-time consistency, and a sample is shown in Fig.~\ref{fig:long_time}. More results can be found \textbf{in the appendix}.
The proposed FreMOTR eliminate irrational temporal inconsistency like the artifacts and noise, and thus the visual quality of individual frames is improved at the same time, which is consistent with the results in Table~\ref{tab:quantitative}.
% Therefore, the visual quality of our results are more natural than C2F-FWN. 

% However, as shown in Table~\ref{tab:quantitative}, FreMOTR has similar or even worse quantitative results such as SSIM, PSNR, and LPIPS. 
% This phenomenon raises another problem that how to evaluate the synthesized results in a more convincing way.

\subsubsection{Improvement of the Temporal Frequency Gap~(EQ3)}
\label{subsubsec:eq3}
\begin{figure}[t]
    \centering
    \includegraphics[width=0.9\columnwidth]{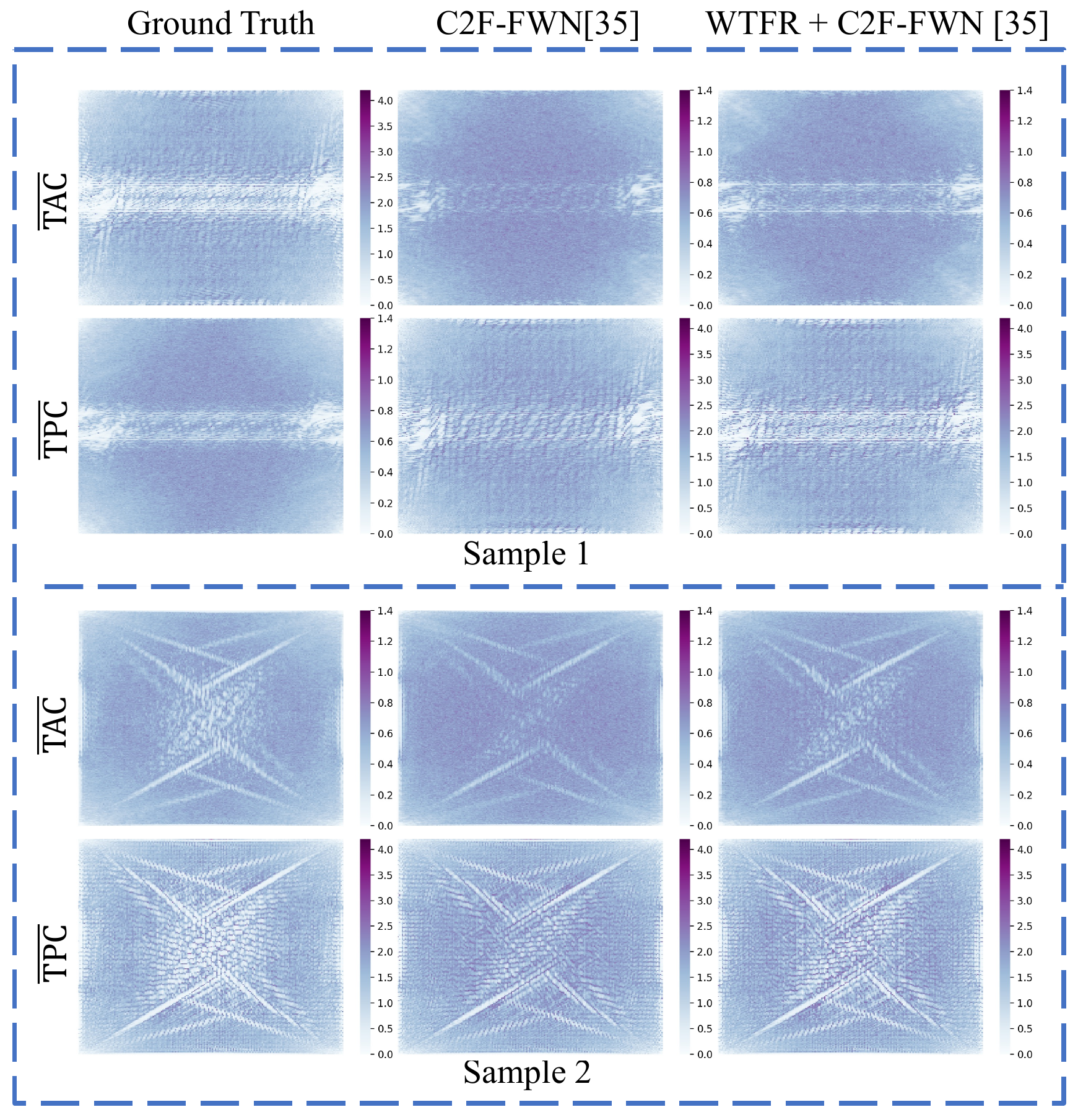}
    \caption{Example heatmaps of mean Temporal Frequency Change $\overline{\mathbf{TFC}}$~(i.e., $\overline{\mathbf{TAC}}$ and $\overline{\mathbf{TPC}}$) from ground truth videos, synthetic videos of original models~(i.e., C2F-FWN~\cite{aaai/WeiXSH2021}), and synthetic videos of WTFR-enhanced models. With the proposed WTFR, the $\overline{\mathbf{TFC}}$ of synthetic videos are closer to thoes of natural videos.}
    \vspace{-5mm}
    \label{fig:case_tfc} 
\end{figure}

%\ \\
We combined the WTFR directly with the backbone to analyze the effectiveness of WTFR~(i.e., only WTFR). 
We visualize the video-level TFC to obtain an observation of overall improvement. 
For a video with $\rm T$ frames, the mean TFC~(i.e., $\overline{\rm TFC})$ for amplitude and phase over the entire video~(i.e., $\overline{\rm TAC}$ and $\overline{\rm TPC}$) are formulated as
\begin{equation}
\begin{aligned}
\label{eq:mean_tfc}
\overline{TAC}(u,v)&=\frac{1}{T-1}\sum_{t=2}^{T}{TAC_t}(u,v),\\
\overline{TPC}(u,v)&=\frac{1}{T-1}\sum_{t=2}^{T}{TPC_t}(u,v),
\end{aligned} \vspace{-2mm}
\end{equation}
where $(u,v)$ is the coordinate on the spectra. 
Several samples are shown in Fig.~\ref{fig:case_tfc}. C2F-FWN considers temporal consistency based on optical flow, and the patterns of $\overline{\rm TFC}$ from C2F-FWN is similar with the ones of ground truth. However, the values are larger in general, especially in the low-frequency components.
Our proposed WTFR can improve the synthetic results, making not only the distribution of $\overline{\rm TAC}$ and $\overline{\rm TPC}$ but also the values closer to natural videos.

The experimental results demonstrate that the proposed frequency-domain regularization~(i.e., WTFR) indeed makes the temporal frequency changes of synthetic videos closer to the ones of natural videos, which component the pixel-domain supervision for human motion transfer.
More comparisons of frames and spectra can be found in the \textbf{appendix materials}.

\subsection{Ablation Study~(EQ4)}
\label{subsec:ablation}

\subsubsection{Ablation of the FreMOTR Components}
% \ \\
\vspace{-1mm}

We first conducted the ablation study on the two components of FreMOTR.
The results are presented in Table~\ref{tab:quantitative}. 
``only FFC'' refers to the model trained from the second stage without WTFR finetuning
``only WTFR'' refers to the backbone model finetuned with WTFR. 

Specifically, the WTFR significantly improves the frame-level image quality, especially the LPIPS and PSNR. 
WTFR improves the TCM a lot at the same time. 
We analyze that WTFR reduce the irrational noise like artifacts, noise, and jitters that suddenly appear, and thus achieves better image quality and temporal consistency.

The FAR significantly improve the TCM. We infer that FAR mainly mitigate the incoherent appearance of the person and achieve much better temporal consistency. 
To make the entire image more natural, the FAR takes the entire image as input rather than only the body region, and the background may be slightly influenced, which lead to unsatisfied frame quality metrics. 
However, by observing the results, we did not observe image quality descending. In contrary, as shown in the qualitative analysis, although FreMOTR and ``only FFC"" have similar or even worse image quality metrics with C2F-FWN, they have less obvious artifacts and unnatural patches.

For human motion transfer, both frame-level image quality and temporal consistency are important, and the Full FreMOTR achieves a well trade-off. The temporal consistency metric of FreMOTR is much better than the backbone and slightly worse than ``only FFC'', while the image quality metrics like LPIPS and PSNR is better than ``only FFC'' and the backbone. It is reasonable that the smooth
Furthermore, the qualitative analysis in Sec.~\ref{subsec:qualitative} demonstrates that beyond the frame-level image quality, the visual performance is significantly improved.
These results demonstrate that all the components of FreMOTR works, which benefit temporally consistent human motion transfer in different aspects.

\begin{table}[t]
\centering
\caption{Ablation study of FreMOTR, with the variations that removes FAR~(i.e., only WTFR) and WTFR~(i.e., only FAR), respectively. TCM is a temporal consistency metric, while LPIPS, PSNR, and SSIM are frame-level image quality metrics.$\uparrow$ means higher is better, while $\downarrow$ means lower is better. The best and second-best results in each column are {boldfaced} and {underlined}, respectively. The proposed FreMOTR achieves the best aggregative results on four metrics.} \vspace{-3mm}
\label{tab:quantitative_ablation}
\begin{tabular}{@{} c l l l l l l @{}}
\toprule
\multirow{2}{*}{Components} & & \multicolumn{1}{c}{{Temporal}} && \multicolumn{3}{c}{Frame Quality}\\
\cmidrule{3-3} \cmidrule{5-7}
&& TCM $\uparrow$ && LPIPS $\downarrow$ &PSNR $\uparrow$ & SSIM $\uparrow$ \\
\midrule
Full FreMOTR & & \underline{0.678} && \underline{0.059} & \underline{22.25} & 0.780 \\
only WTFR & & 0.592 && \textbf{0.052} & \textbf{22.86} & \textbf{0.839} \\
only FAR & & \textbf{0.681} && 0.060 & 22.11 & 0.780 \\
backbone & & {0.523} && {0.060} & {22.17} & \underline{0.821} \\
\bottomrule
\end{tabular}
\end{table}

\subsubsection{Ablation of the Frequency Spectrum}
WTFR is composed with amplitude and phase. To study the roles of each components, we combined the converged backbone~(i.e., C2F-FWN~\cite{aaai/WeiXSH2021}) with several variation of WTFR, and the results are presented in Table~\ref{tab:ablation}. 
The WTFR is defined in Eq.~\ref{eq:L_TAC}, Eq.~\ref{eq:L_TPC}, and Eq.~\ref{eq:wtfr}. In the table, ``Phase'' refers to phase only~(i.e., $\alpha=0$ in Eq.~\ref{eq:wtfr}), ``Amplitude'' refers to amplitude only~(i.e., $\beta=0$ in Eq.~\ref{eq:wtfr}). ``TFR'' means removing the weighted term $w$~(i.e., $w=1$ in Eq.~\ref{eq:L_TAC} and Eq.~\ref{eq:L_TPC}). For the Mixed setting, only one of the components uses the weighted term. ``AMP.+WP'' means amplitude without the weighted term $w$ and phase with the weighted term $w$, 
while ``PH.+WA'' means phase without the weighted term $w$ and amplitude with the weighted term $w$.

First, the results demonstrate that all variations improve the temporal consistency, while the frame-level quality can also be improved. The amplitude part and phase part of WTFR are both effective.
The TCM metrics of generated videos are improved by {11\% $\sim$ 14\%} than the original models.

Second, the amplitude part is more important than the phase. As described in Sec.~\ref{subsec:DFT}, amplitude reflects the intensity of the pixels while phase reflects the position. 
Therefore, minimizing amplitude fluctuation can guarantee the consistency of pixel intensity, which reduces irrational components that suddenly appear.

Third, the comparison between results of WTFR, TFR, and Mixed shows that the mixed variation of ``PH. + WA'' works best. The weighting term in WTFR is important for the model to pay more attention to high-frequency components, especially for the amplitude. This phenomenon is consistent with the observation in Fig.~\ref{fig:tfc}. The main content~(i.e., intensity of the pixels) of adjacent frames does not change too much, and thus we should adopt larger penalty to the high-frequency changes for amplitude spectrum to make the synthetic frames more natural. On the other hand, the position of the pixels change but not very intensely, and thus we just adopt a uniform penalty for all components in the phase spectrum.

\begin{table}[t]
\setlength{\belowcaptionskip}{-7mm}
\centering
\caption{Ablation study of WTFR. Temp. is shorten for Temporal. TCM is a temporal consistency metric, while LPIPS, PSNR, and SSIM are frame-level metrics. $\uparrow$ means higher is better, while $\downarrow$ means lower is better. The best and second-best results are {boldfaced} and {underlined}, respectively.}  \vspace{-3mm}
\label{tab:ablation}
\begin{tabular}{@{} c l l l l l l l l l l @{}}
\toprule
\multirow{2}{*}{TFR Type} & \multirow{2}{*}{Setting} & \multicolumn{1}{c}{Temp.} && \multicolumn{3}{c}{Frame Quality}\\
\cmidrule{3-3} \cmidrule{5-7}
&& {TCM} $\uparrow$ && LPIPS $\downarrow$ & PSNR $\uparrow$ & SSIM $\uparrow$ \\
\midrule
backbone & --- & 0.523 && 0.060 & 22.172 & 0.821 \\
\midrule 
\multirow{3}{*}{TFR} & Phase & 0.590 && 0.056 & 22.535 & 0.833 \\
& Amplitude & 0.589 && 0.054 & 22.652 & 0.833 \\
& Both & 0.583 && 0.052 & 22.772 & 0.836 \\
\midrule 
\multirow{3}{*}{WTFR} & Phase & 0.584 && 0.053 & 22.797 & 0.838 \\
& Amplitude & \underline{0.591} && \underline{0.052} & \textbf{22.860} & 0.838 \\
& Both & 0.591 && 0.053 & 22.846 & 0.838 \\
\midrule 
\multirow{2}{*}{Mixed} & AMP.+WP & 0.591 && 0.052 & 22.820 & 0.838 \\
& PH.+WA & \textbf{0.592} && \textbf{0.052} & \underline{22.857} & \textbf{0.839} \\
\bottomrule
\end{tabular} \vspace{-7mm}
\end{table}

% \vspace{-3mm}

\section{Conclusion}
\vspace{-1mm}
In this paper, we delve into the frequency space for temporally consistent human motion transfer. We make the first comprehensive analysis of natural and synthetic videos in the frequency domain to reveal the frequency gap.
% in both the spatial domain of individual frames and the temporal dimension of the video. 
We propose a novel \textbf{FreMOTR} framework that can effectively mitigate the spatial artifacts and the temporal inconsistency of the generated videos with two novel frequency-based regularization modules. Experiments results demonstrate that the FreMOTR not only yields superior performance in temporal consistency metrics but also improves the frame-level visual quality of synthetic videos.
In particular, the temporal consistency metrics are improved by nearly 30\% than the state-of-the-art models.

\begin{acks}
\vspace{-1mm}
This work was supported by the National Key R\&D Program of China (No.2020AAA0103800), the Zhejiang Provincial Key R\&D Program of China (No. 2021C01164), and the Project of Chinese Academy of Sciences (E141020). Juan Cao thanks the Nanjing Government Affairs and Public Opinion Research Institute for the support of "CaoJuan Studio" and thanks Chi Peng, Jingjing Jiang, Qiang Liu, and Yu Dai for their help.
\end{acks}

\balance
\bibliographystyle{ACM-Reference-Format}
\bibliography{camera_ready_acm}


\begin{thebibliography}{44}

%%% the Web address.

\ifx \showCODEN    \undefined \def \showCODEN     #1{\unskip}     \fi
\ifx \showDOI      \undefined \def \showDOI       #1{#1}\fi
\ifx \showISBNx    \undefined \def \showISBNx     #1{\unskip}     \fi
\ifx \showISBNxiii \undefined \def \showISBNxiii  #1{\unskip}     \fi
\ifx \showISSN     \undefined \def \showISSN      #1{\unskip}     \fi
\ifx \showLCCN     \undefined \def \showLCCN      #1{\unskip}     \fi
\ifx \shownote     \undefined \def \shownote      #1{#1}          \fi
\ifx \showarticletitle \undefined \def \showarticletitle #1{#1}   \fi
\ifx \showURL      \undefined \def \showURL       {\relax}        \fi
% invisible to TeX
\providecommand\bibfield[2]{#2}
\providecommand\bibinfo[2]{#2}
\providecommand\natexlab[1]{#1}
\providecommand\showeprint[2][]{arXiv:#2}

\bibitem[Adam et~al\mbox{.}(2019)]%
        {nips/paszke2019}
\bibfield{author}{\bibinfo{person}{Paszke Adam}, \bibinfo{person}{Gross Sam},
  \bibinfo{person}{Francisco Mass}, \bibinfo{person}{Lerer Adam},
  \bibinfo{person}{Bradbury James}, \bibinfo{person}{Chanan Gregory},
  \bibinfo{person}{Killeen Trevor}, \bibinfo{person}{Lin Zeming},
  \bibinfo{person}{Gimelshein Natalia}, \bibinfo{person}{Antiga Luca},
  {et~al\mbox{.}}} \bibinfo{year}{2019}\natexlab{}.
\newblock \showarticletitle{PyTorch: An Imperative Style, High-Performance Deep
  Learning Library}. In \bibinfo{booktitle}{\emph{Advances in Neural
  Information Processing Systems}}. \bibinfo{pages}{8026--8037}.
\newblock


\bibitem[Balakrishnan et~al\mbox{.}(2018)]%
        {cvpr/BalakrishnanZDD18}
\bibfield{author}{\bibinfo{person}{Guha Balakrishnan}, \bibinfo{person}{Amy
  Zhao}, \bibinfo{person}{Adrian~V. Dalca}, \bibinfo{person}{Fr{\'{e}}do
  Durand}, {and} \bibinfo{person}{John~V. Guttag}.}
  \bibinfo{year}{2018}\natexlab{}.
\newblock \showarticletitle{Synthesizing Images of Humans in Unseen Poses}. In
  \bibinfo{booktitle}{\emph{{IEEE} Conference on Computer Vision and Pattern
  Recognition}}. \bibinfo{pages}{8340--8348}.
\newblock


\bibitem[Bhattacharjee and Das(2017)]%
        {nips/BhattacharjeeD17}
\bibfield{author}{\bibinfo{person}{Prateep Bhattacharjee} {and}
  \bibinfo{person}{Sukhendu Das}.} \bibinfo{year}{2017}\natexlab{}.
\newblock \showarticletitle{Temporal Coherency based Criteria for Predicting
  Video Frames using Deep Multi-stage Generative Adversarial Networks}. In
  \bibinfo{booktitle}{\emph{Advances in Neural Information Processing
  Systems}}. \bibinfo{pages}{4268--4277}.
\newblock


\bibitem[Bonneel et~al\mbox{.}(2015)]%
        {tog/BonneelTSSPP15}
\bibfield{author}{\bibinfo{person}{Nicolas Bonneel}, \bibinfo{person}{James
  Tompkin}, \bibinfo{person}{Kalyan Sunkavalli}, \bibinfo{person}{Deqing Sun},
  \bibinfo{person}{Sylvain Paris}, {and} \bibinfo{person}{Hanspeter Pfister}.}
  \bibinfo{year}{2015}\natexlab{}.
\newblock \showarticletitle{Blind video temporal consistency}.
\newblock \bibinfo{journal}{\emph{{ACM} Trans. Graph.}} \bibinfo{volume}{34},
  \bibinfo{number}{6} (\bibinfo{year}{2015}), \bibinfo{pages}{196:1--196:9}.
\newblock


\bibitem[Chan et~al\mbox{.}(2019)]%
        {iccv/ChanGZE19}
\bibfield{author}{\bibinfo{person}{Caroline Chan}, \bibinfo{person}{Shiry
  Ginosar}, \bibinfo{person}{Tinghui Zhou}, {and} \bibinfo{person}{Alexei~A.
  Efros}.} \bibinfo{year}{2019}\natexlab{}.
\newblock \showarticletitle{Everybody Dance Now}. In
  \bibinfo{booktitle}{\emph{{IEEE} International Conference on Computer
  Vision}}. \bibinfo{pages}{5932--5941}.
\newblock


\bibitem[Chen et~al\mbox{.}(2021)]%
        {aaai/ChenLJLL2021}
\bibfield{author}{\bibinfo{person}{Yuanqi Chen}, \bibinfo{person}{Ge Li},
  \bibinfo{person}{Cece Jin}, \bibinfo{person}{Shan Liu}, {and}
  \bibinfo{person}{Thomas Li}.} \bibinfo{year}{2021}\natexlab{}.
\newblock \showarticletitle{{SSD-GAN:} Measuring the Realness in the Spatial
  and Spectral Domains}. In \bibinfo{booktitle}{\emph{AAAI Conference on
  Artificial Intelligence}}.
\newblock


\bibitem[Chi et~al\mbox{.}(2020)]%
        {nips/chi2020}
\bibfield{author}{\bibinfo{person}{Lu Chi}, \bibinfo{person}{Borui Jiang},
  {and} \bibinfo{person}{Yadong Mu}.} \bibinfo{year}{2020}\natexlab{}.
\newblock \showarticletitle{Fast fourier convolution}.
\newblock \bibinfo{journal}{\emph{Advances in Neural Information Processing
  Systems}}  \bibinfo{volume}{33} (\bibinfo{year}{2020}),
  \bibinfo{pages}{4479--4488}.
\newblock


\bibitem[Dong et~al\mbox{.}(2018)]%
        {nips/DongLGLZY18}
\bibfield{author}{\bibinfo{person}{Haoye Dong}, \bibinfo{person}{Xiaodan
  Liang}, \bibinfo{person}{Ke Gong}, \bibinfo{person}{Hanjiang Lai},
  \bibinfo{person}{Jia Zhu}, {and} \bibinfo{person}{Jian Yin}.}
  \bibinfo{year}{2018}\natexlab{}.
\newblock \showarticletitle{Soft-Gated Warping-{GAN} for Pose-Guided Person
  Image Synthesis}. In \bibinfo{booktitle}{\emph{Advances in Neural Information
  Processing Systems}}. \bibinfo{pages}{472--482}.
\newblock


\bibitem[Durall et~al\mbox{.}(2020)]%
        {cvpr/DurallKK20}
\bibfield{author}{\bibinfo{person}{Ricard Durall}, \bibinfo{person}{Margret
  Keuper}, {and} \bibinfo{person}{Janis Keuper}.}
  \bibinfo{year}{2020}\natexlab{}.
\newblock \showarticletitle{Watch Your Up-Convolution: {CNN} Based Generative
  Deep Neural Networks Are Failing to Reproduce Spectral Distributions}. In
  \bibinfo{booktitle}{\emph{{IEEE} Conference on Computer Vision and Pattern
  Recognition}}. \bibinfo{pages}{7887--7896}.
\newblock


\bibitem[Dzanic et~al\mbox{.}(2020)]%
        {nips/DzanicSW20}
\bibfield{author}{\bibinfo{person}{Tarik Dzanic}, \bibinfo{person}{Karan Shah},
  {and} \bibinfo{person}{Freddie~D. Witherden}.}
  \bibinfo{year}{2020}\natexlab{}.
\newblock \showarticletitle{Fourier Spectrum Discrepancies in Deep Network
  Generated Images}. In \bibinfo{booktitle}{\emph{Advances in Neural
  Information Processing Systems}}.
\newblock


\bibitem[Efros et~al\mbox{.}(2003)]%
        {iccv/EfrosBMM03}
\bibfield{author}{\bibinfo{person}{Alexei~A. Efros},
  \bibinfo{person}{Alexander~C. Berg}, \bibinfo{person}{Greg Mori}, {and}
  \bibinfo{person}{Jitendra Malik}.} \bibinfo{year}{2003}\natexlab{}.
\newblock \showarticletitle{Recognizing Action at a Distance}. In
  \bibinfo{booktitle}{\emph{{IEEE} International Conference on Computer
  Vision}}. \bibinfo{pages}{726--733}.
\newblock


\bibitem[Frank et~al\mbox{.}(2020)]%
        {icml/FrankESFKH20}
\bibfield{author}{\bibinfo{person}{Joel Frank}, \bibinfo{person}{Thorsten
  Eisenhofer}, \bibinfo{person}{Lea Sch{\"{o}}nherr}, \bibinfo{person}{Asja
  Fischer}, \bibinfo{person}{Dorothea Kolossa}, {and} \bibinfo{person}{Thorsten
  Holz}.} \bibinfo{year}{2020}\natexlab{}.
\newblock \showarticletitle{Leveraging Frequency Analysis for Deep Fake Image
  Recognition}. In \bibinfo{booktitle}{\emph{International Conference on
  Machine Learning}}, Vol.~\bibinfo{volume}{119}. \bibinfo{pages}{3247--3258}.
\newblock


\bibitem[Goodfellow et~al\mbox{.}(2014)]%
        {nips/GoodfellowPMXWOCB14}
\bibfield{author}{\bibinfo{person}{Ian~J. Goodfellow}, \bibinfo{person}{Jean
  Pouget{-}Abadie}, \bibinfo{person}{Mehdi Mirza}, \bibinfo{person}{Bing Xu},
  \bibinfo{person}{David Warde{-}Farley}, \bibinfo{person}{Sherjil Ozair},
  \bibinfo{person}{Aaron~C. Courville}, {and} \bibinfo{person}{Yoshua Bengio}.}
  \bibinfo{year}{2014}\natexlab{}.
\newblock \showarticletitle{Generative Adversarial Nets}. In
  \bibinfo{booktitle}{\emph{Advances in Neural Information Processing
  Systems}}. \bibinfo{pages}{2672--2680}.
\newblock


\bibitem[Han et~al\mbox{.}(2019)]%
        {iccv/han2019}
\bibfield{author}{\bibinfo{person}{Xintong Han}, \bibinfo{person}{Xiaojun Hu},
  \bibinfo{person}{Weilin Huang}, {and} \bibinfo{person}{Matthew~R Scott}.}
  \bibinfo{year}{2019}\natexlab{}.
\newblock \showarticletitle{Clothflow: A flow-based model for clothed person
  generation}. In \bibinfo{booktitle}{\emph{{IEEE/CVF} International Conference
  on Computer Vision}}. \bibinfo{pages}{10471--10480}.
\newblock


\bibitem[Ilg et~al\mbox{.}(2017)]%
        {cvpr/ilg2017}
\bibfield{author}{\bibinfo{person}{Eddy Ilg}, \bibinfo{person}{Mayer Nikolaus},
  \bibinfo{person}{Saikia Tonmoy}, \bibinfo{person}{Keuper Margret},
  \bibinfo{person}{Dosovitskiy Alexey}, {and} \bibinfo{person}{Brox Thomas}.}
  \bibinfo{year}{2017}\natexlab{}.
\newblock \showarticletitle{Flownet 2.0: Evolution of optical flow estimation
  with deep networks}. In \bibinfo{booktitle}{\emph{{IEEE} Conference on
  Computer Vision and Pattern Recognition}}. \bibinfo{pages}{2462--2470}.
\newblock


\bibitem[Johnson et~al\mbox{.}(2016)]%
        {eccv/Johnson16}
\bibfield{author}{\bibinfo{person}{Justin Johnson}, \bibinfo{person}{Alexandre
  Alahi}, {and} \bibinfo{person}{Li Fei-Fei}.} \bibinfo{year}{2016}\natexlab{}.
\newblock \showarticletitle{Perceptual losses for real-time style transfer and
  super-resolution}. In \bibinfo{booktitle}{\emph{European Conference on
  Computer Vision}}. \bibinfo{pages}{694--711}.
\newblock


\bibitem[Kingma and Ba(2015)]%
        {iclr/KingmaB14}
\bibfield{author}{\bibinfo{person}{Diederik~P. Kingma} {and}
  \bibinfo{person}{Jimmy Ba}.} \bibinfo{year}{2015}\natexlab{}.
\newblock \showarticletitle{Adam: {A} Method for Stochastic Optimization}. In
  \bibinfo{booktitle}{\emph{International Conference on Learning
  Representations}}.
\newblock


\bibitem[Lai et~al\mbox{.}(2018)]%
        {eccv/LaiHWSYY18}
\bibfield{author}{\bibinfo{person}{Wei{-}Sheng Lai}, \bibinfo{person}{Jia{-}Bin
  Huang}, \bibinfo{person}{Oliver Wang}, \bibinfo{person}{Eli Shechtman},
  \bibinfo{person}{Ersin Yumer}, {and} \bibinfo{person}{Ming{-}Hsuan Yang}.}
  \bibinfo{year}{2018}\natexlab{}.
\newblock \showarticletitle{Learning Blind Video Temporal Consistency}. In
  \bibinfo{booktitle}{\emph{European Conference on Computer Vision}},
  Vol.~\bibinfo{volume}{11219}. \bibinfo{pages}{179--195}.
\newblock


\bibitem[Lei et~al\mbox{.}(2020)]%
        {nips/LeiXC20}
\bibfield{author}{\bibinfo{person}{Chenyang Lei}, \bibinfo{person}{Yazhou
  Xing}, {and} \bibinfo{person}{Qifeng Chen}.} \bibinfo{year}{2020}\natexlab{}.
\newblock \showarticletitle{Blind Video Temporal Consistency via Deep Video
  Prior}. In \bibinfo{booktitle}{\emph{Advances in Neural Information
  Processing Systems}}.
\newblock


\bibitem[Liu et~al\mbox{.}(2018)]%
        {aaai/liu2018t}
\bibfield{author}{\bibinfo{person}{Kun Liu}, \bibinfo{person}{Wu Liu},
  \bibinfo{person}{Chuang Gan}, \bibinfo{person}{Mingkui Tan}, {and}
  \bibinfo{person}{Huadong Ma}.} \bibinfo{year}{2018}\natexlab{}.
\newblock \showarticletitle{T-C3D: Temporal convolutional 3D network for
  real-time action recognition}. In \bibinfo{booktitle}{\emph{Proceedings of
  the AAAI conference on artificial intelligence}}, Vol.~\bibinfo{volume}{32}.
\newblock


\bibitem[Liu et~al\mbox{.}(2019c)]%
        {tog/LiuXZKBHWT19}
\bibfield{author}{\bibinfo{person}{Lingjie Liu}, \bibinfo{person}{Weipeng Xu},
  \bibinfo{person}{Michael Zollh{\"{o}}fer}, \bibinfo{person}{Hyeongwoo Kim},
  \bibinfo{person}{Florian Bernard}, \bibinfo{person}{Marc Habermann},
  \bibinfo{person}{Wenping Wang}, {and} \bibinfo{person}{Christian Theobalt}.}
  \bibinfo{year}{2019}\natexlab{c}.
\newblock \showarticletitle{Neural Rendering and Reenactment of Human Actor
  Videos}.
\newblock \bibinfo{journal}{\emph{{ACM} Trans. Graph.}} \bibinfo{volume}{38},
  \bibinfo{number}{5} (\bibinfo{year}{2019}), \bibinfo{pages}{139:1--139:14}.
\newblock


\bibitem[Liu et~al\mbox{.}(2022)]%
        {liu2021recent}
\bibfield{author}{\bibinfo{person}{Wu Liu}, \bibinfo{person}{Qian Bao},
  \bibinfo{person}{Yu Sun}, {and} \bibinfo{person}{Tao Mei}.}
  \bibinfo{year}{2022}\natexlab{}.
\newblock \showarticletitle{Recent advances in monocular 2d and 3d human pose
  estimation: A deep learning perspective}.
\newblock \bibinfo{journal}{\emph{Comput. Surveys}} (\bibinfo{year}{2022}).
\newblock


\bibitem[Liu et~al\mbox{.}(2019b)]%
        {iccv/LiuPML0G19}
\bibfield{author}{\bibinfo{person}{Wen Liu}, \bibinfo{person}{Zhixin Piao},
  \bibinfo{person}{Jie Min}, \bibinfo{person}{Wenhan Luo}, \bibinfo{person}{Lin
  Ma}, {and} \bibinfo{person}{Shenghua Gao}.} \bibinfo{year}{2019}\natexlab{b}.
\newblock \showarticletitle{Liquid Warping {GAN:} {A} Unified Framework for
  Human Motion Imitation, Appearance Transfer and Novel View Synthesis}. In
  \bibinfo{booktitle}{\emph{{IEEE} International Conference on Computer
  Vision}}. \bibinfo{pages}{5903--5912}.
\newblock


\bibitem[Liu et~al\mbox{.}(2021)]%
        {tpami/LiuPTLMG2021}
\bibfield{author}{\bibinfo{person}{Wen Liu}, \bibinfo{person}{Zhixin Piao},
  \bibinfo{person}{Zhi Tu}, \bibinfo{person}{Wenhan Luo}, \bibinfo{person}{Lin
  Ma}, {and} \bibinfo{person}{Shenghua Gao}.} \bibinfo{year}{2021}\natexlab{}.
\newblock \showarticletitle{Liquid Warping {GAN} with Attention: {A} Unified
  Framework for Human Image Synthesis}.
\newblock \bibinfo{journal}{\emph{{IEEE} Trans. Pattern Anal. Mach. Intell.}}
  (\bibinfo{year}{2021}).
\newblock


\bibitem[Liu et~al\mbox{.}(2019a)]%
        {cvpr/LiuLZCGYM19}
\bibfield{author}{\bibinfo{person}{Xinchen Liu}, \bibinfo{person}{Wu Liu},
  \bibinfo{person}{Meng Zhang}, \bibinfo{person}{Jingwen Chen},
  \bibinfo{person}{Lianli Gao}, \bibinfo{person}{Chenggang Yan}, {and}
  \bibinfo{person}{Tao Mei}.} \bibinfo{year}{2019}\natexlab{a}.
\newblock \showarticletitle{Social Relation Recognition From Videos via
  Multi-Scale Spatial-Temporal Reasoning}. In \bibinfo{booktitle}{\emph{{IEEE}
  Conference on Computer Vision and Pattern Recognition}}.
  \bibinfo{pages}{3566--3574}.
\newblock


\bibitem[Liu et~al\mbox{.}(2019d)]%
        {mm/LiuZLSM19}
\bibfield{author}{\bibinfo{person}{Xinchen Liu}, \bibinfo{person}{Meng Zhang},
  \bibinfo{person}{Wu Liu}, \bibinfo{person}{Jingkuan Song}, {and}
  \bibinfo{person}{Tao Mei}.} \bibinfo{year}{2019}\natexlab{d}.
\newblock \showarticletitle{BraidNet: Braiding Semantics and Details for
  Accurate Human Parsing}. In \bibinfo{booktitle}{\emph{ACM Conference on
  Multimedia}}, \bibfield{editor}{\bibinfo{person}{Laurent Amsaleg},
  \bibinfo{person}{Benoit Huet}, \bibinfo{person}{Martha~A. Larson},
  \bibinfo{person}{Guillaume Gravier}, \bibinfo{person}{Hayley Hung},
  \bibinfo{person}{Chong{-}Wah Ngo}, {and} \bibinfo{person}{Wei~Tsang Ooi}}
  (Eds.). \bibinfo{pages}{338--346}.
\newblock


\bibitem[Rahaman et~al\mbox{.}(2019)]%
        {icml/RahamanBADLHBC19}
\bibfield{author}{\bibinfo{person}{Nasim Rahaman}, \bibinfo{person}{Aristide
  Baratin}, \bibinfo{person}{Devansh Arpit}, \bibinfo{person}{Felix Draxler},
  \bibinfo{person}{Min Lin}, \bibinfo{person}{Fred~A. Hamprecht},
  \bibinfo{person}{Yoshua Bengio}, {and} \bibinfo{person}{Aaron~C. Courville}.}
  \bibinfo{year}{2019}\natexlab{}.
\newblock \showarticletitle{On the Spectral Bias of Neural Networks}. In
  \bibinfo{booktitle}{\emph{International Conference on Machine Learning}},
  Vol.~\bibinfo{volume}{97}. \bibinfo{pages}{5301--5310}.
\newblock


\bibitem[Richard et~al\mbox{.}(2018)]%
        {cvpr/zhang2018}
\bibfield{author}{\bibinfo{person}{Zhang Richard}, \bibinfo{person}{Isola
  Phillip}, \bibinfo{person}{Efros~Alexei A}, \bibinfo{person}{Shechtman Eli},
  {and} \bibinfo{person}{Oliver Wang}.} \bibinfo{year}{2018}\natexlab{}.
\newblock \showarticletitle{The unreasonable effectiveness of deep features as
  a perceptual metric}. In \bibinfo{booktitle}{\emph{{IEEE} Conference on
  Computer Vision and Pattern Recognition}}. \bibinfo{pages}{586--595}.
\newblock


\bibitem[Seshadrinathan and Bovik(2010)]%
        {tip/SeshadrinathanB10}
\bibfield{author}{\bibinfo{person}{Kalpana Seshadrinathan} {and}
  \bibinfo{person}{Alan~C. Bovik}.} \bibinfo{year}{2010}\natexlab{}.
\newblock \showarticletitle{Motion Tuned Spatio-Temporal Quality Assessment of
  Natural Videos}.
\newblock \bibinfo{journal}{\emph{{IEEE} Trans. Image Process.}}
  \bibinfo{volume}{19}, \bibinfo{number}{2} (\bibinfo{year}{2010}),
  \bibinfo{pages}{335--350}.
\newblock


\bibitem[Siarohin et~al\mbox{.}(2019)]%
        {cvpr/SiarohinLT0S19}
\bibfield{author}{\bibinfo{person}{Aliaksandr Siarohin},
  \bibinfo{person}{St{\'{e}}phane Lathuili{\`{e}}re}, \bibinfo{person}{Sergey
  Tulyakov}, \bibinfo{person}{Elisa Ricci}, {and} \bibinfo{person}{Nicu Sebe}.}
  \bibinfo{year}{2019}\natexlab{}.
\newblock \showarticletitle{Animating Arbitrary Objects via Deep Motion
  Transfer}. In \bibinfo{booktitle}{\emph{{IEEE} Conference on Computer Vision
  and Pattern Recognition}}. \bibinfo{pages}{2377--2386}.
\newblock


\bibitem[Sun et~al\mbox{.}(2022)]%
        {sun2022putting}
\bibfield{author}{\bibinfo{person}{Yu Sun}, \bibinfo{person}{Wu Liu},
  \bibinfo{person}{Qian Bao}, \bibinfo{person}{Yili Fu}, \bibinfo{person}{Tao
  Mei}, {and} \bibinfo{person}{Michael~J Black}.}
  \bibinfo{year}{2022}\natexlab{}.
\newblock \showarticletitle{Putting people in their place: Monocular regression
  of 3d people in depth}. In \bibinfo{booktitle}{\emph{{IEEE/CVF} Conference on
  Computer Vision and Pattern Recognition}}. \bibinfo{pages}{13243--13252}.
\newblock


\bibitem[Suvorov et~al\mbox{.}(2022)]%
        {wacv/Suvorov2022}
\bibfield{author}{\bibinfo{person}{Roman Suvorov}, \bibinfo{person}{Elizaveta
  Logacheva}, \bibinfo{person}{Anton Mashikhin}, \bibinfo{person}{Anastasia
  Remizova}, \bibinfo{person}{Arsenii Ashukha}, \bibinfo{person}{Aleksei
  Silvestrov}, \bibinfo{person}{Naejin Kong}, \bibinfo{person}{Harshith Goka},
  \bibinfo{person}{Kiwoong Park}, {and} \bibinfo{person}{Victor Lempitsky}.}
  \bibinfo{year}{2022}\natexlab{}.
\newblock \showarticletitle{Resolution-robust Large Mask Inpainting with
  Fourier Convolutions}. In \bibinfo{booktitle}{\emph{{IEEE/CVF} Winter
  Conference on Applications of Computer Vision}}. \bibinfo{pages}{2149--2159}.
\newblock


\bibitem[Tancik et~al\mbox{.}(2020)]%
        {nips/TancikSMFRSRBN20}
\bibfield{author}{\bibinfo{person}{Matthew Tancik}, \bibinfo{person}{Pratul~P.
  Srinivasan}, \bibinfo{person}{Ben Mildenhall}, \bibinfo{person}{Sara
  Fridovich{-}Keil}, \bibinfo{person}{Nithin Raghavan},
  \bibinfo{person}{Utkarsh Singhal}, \bibinfo{person}{Ravi Ramamoorthi},
  \bibinfo{person}{Jonathan~T. Barron}, {and} \bibinfo{person}{Ren Ng}.}
  \bibinfo{year}{2020}\natexlab{}.
\newblock \showarticletitle{Fourier Features Let Networks Learn High Frequency
  Functions in Low Dimensional Domains}. In \bibinfo{booktitle}{\emph{Advances
  in Neural Information Processing Systems}}.
\newblock


\bibitem[Thies et~al\mbox{.}(2020)]%
        {eccv/ThiesETTN20}
\bibfield{author}{\bibinfo{person}{Justus Thies}, \bibinfo{person}{Mohamed
  Elgharib}, \bibinfo{person}{Ayush Tewari}, \bibinfo{person}{Christian
  Theobalt}, {and} \bibinfo{person}{Matthias Nie{\ss}ner}.}
  \bibinfo{year}{2020}\natexlab{}.
\newblock \showarticletitle{Neural Voice Puppetry: Audio-Driven Facial
  Reenactment}. In \bibinfo{booktitle}{\emph{European Conference on Computer
  Vision}}, Vol.~\bibinfo{volume}{12361}. \bibinfo{pages}{716--731}.
\newblock


\bibitem[Tulyakov et~al\mbox{.}(2018)]%
        {cvpr/Tulyakov0YK18}
\bibfield{author}{\bibinfo{person}{Sergey Tulyakov}, \bibinfo{person}{Ming{-}Yu
  Liu}, \bibinfo{person}{Xiaodong Yang}, {and} \bibinfo{person}{Jan Kautz}.}
  \bibinfo{year}{2018}\natexlab{}.
\newblock \showarticletitle{MoCoGAN: Decomposing Motion and Content for Video
  Generation}. In \bibinfo{booktitle}{\emph{{IEEE} Conference on Computer
  Vision and Pattern Recognition}}. \bibinfo{pages}{1526--1535}.
\newblock


\bibitem[Wang et~al\mbox{.}(2019)]%
        {nips/Wang0TLCK19}
\bibfield{author}{\bibinfo{person}{Ting{-}Chun Wang},
  \bibinfo{person}{Ming{-}Yu Liu}, \bibinfo{person}{Andrew Tao},
  \bibinfo{person}{Guilin Liu}, \bibinfo{person}{Bryan Catanzaro}, {and}
  \bibinfo{person}{Jan Kautz}.} \bibinfo{year}{2019}\natexlab{}.
\newblock \showarticletitle{Few-shot Video-to-Video Synthesis}. In
  \bibinfo{booktitle}{\emph{Advances in Neural Information Processing
  Systems}}. \bibinfo{pages}{5014--5025}.
\newblock


\bibitem[Wang et~al\mbox{.}(2018)]%
        {nips/Wang0ZYTKC18}
\bibfield{author}{\bibinfo{person}{Ting{-}Chun Wang},
  \bibinfo{person}{Ming{-}Yu Liu}, \bibinfo{person}{Jun{-}Yan Zhu},
  \bibinfo{person}{Nikolai Yakovenko}, \bibinfo{person}{Andrew Tao},
  \bibinfo{person}{Jan Kautz}, {and} \bibinfo{person}{Bryan Catanzaro}.}
  \bibinfo{year}{2018}\natexlab{}.
\newblock \showarticletitle{Video-to-Video Synthesis}. In
  \bibinfo{booktitle}{\emph{Advances in Neural Information Processing
  Systems}}. \bibinfo{pages}{1152--1164}.
\newblock


\bibitem[Wang et~al\mbox{.}(2004)]%
        {tip/wang04}
\bibfield{author}{\bibinfo{person}{Zhou Wang}, \bibinfo{person}{Bovik~Alan C},
  \bibinfo{person}{Sheikh~Hamid R}, {and} \bibinfo{person}{Simoncelli~Eero P}.}
  \bibinfo{year}{2004}\natexlab{}.
\newblock \showarticletitle{Image quality assessment: from error visibility to
  structural similarity}.
\newblock \bibinfo{journal}{\emph{{IEEE} Trans. Image Process.}}
  \bibinfo{volume}{13}, \bibinfo{number}{4} (\bibinfo{year}{2004}),
  \bibinfo{pages}{600--612}.
\newblock


\bibitem[Wei et~al\mbox{.}(2021a)]%
        {tmm/WeiSH2020}
\bibfield{author}{\bibinfo{person}{Dongxu Wei}, \bibinfo{person}{Haibin Shen},
  {and} \bibinfo{person}{Kejie Huang}.} \bibinfo{year}{2021}\natexlab{a}.
\newblock \showarticletitle{{GAC-GAN}: A General Method for
  Appearance-Controllable Human Video Motion Transfer}.
\newblock \bibinfo{journal}{\emph{{IEEE} Trans. Multim.}}  \bibinfo{volume}{23}
  (\bibinfo{year}{2021}), \bibinfo{pages}{2457--2470}.
\newblock


\bibitem[Wei et~al\mbox{.}(2021b)]%
        {aaai/WeiXSH2021}
\bibfield{author}{\bibinfo{person}{Dongxu Wei}, \bibinfo{person}{Xiaowei Xu},
  \bibinfo{person}{Haibin Shen}, {and} \bibinfo{person}{Kejie Huang}.}
  \bibinfo{year}{2021}\natexlab{b}.
\newblock \showarticletitle{{C2F-FWN:} Coarse-to-Fine Flow Warping Network for
  Spatial-Temporal Consistent Motion Transfer}. In
  \bibinfo{booktitle}{\emph{AAAI Conference on Artificial Intelligence}}.
  \bibinfo{pages}{2852--2860}.
\newblock


\bibitem[Xu et~al\mbox{.}(2011)]%
        {tog/XuLSTBDSKT11}
\bibfield{author}{\bibinfo{person}{Feng Xu}, \bibinfo{person}{Yebin Liu},
  \bibinfo{person}{Carsten Stoll}, \bibinfo{person}{James Tompkin},
  \bibinfo{person}{Gaurav Bharaj}, \bibinfo{person}{Qionghai Dai},
  \bibinfo{person}{Hans{-}Peter Seidel}, \bibinfo{person}{Jan Kautz}, {and}
  \bibinfo{person}{Christian Theobalt}.} \bibinfo{year}{2011}\natexlab{}.
\newblock \showarticletitle{Video-based characters: creating new human
  performances from a multi-view video database}.
\newblock \bibinfo{journal}{\emph{{ACM} Trans. Graph.}} \bibinfo{volume}{30},
  \bibinfo{number}{4} (\bibinfo{year}{2011}), \bibinfo{pages}{32}.
\newblock


\bibitem[Yao et~al\mbox{.}(2017)]%
        {mm/YaoCC17}
\bibfield{author}{\bibinfo{person}{Chun{-}Han Yao},
  \bibinfo{person}{Chia{-}Yang Chang}, {and} \bibinfo{person}{Shao{-}Yi
  Chien}.} \bibinfo{year}{2017}\natexlab{}.
\newblock \showarticletitle{Occlusion-aware Video Temporal Consistency}. In
  \bibinfo{booktitle}{\emph{{ACM} International Conference on Multimedia}}.
  \bibinfo{pages}{777--785}.
\newblock


\bibitem[Zhang et~al\mbox{.}(2019)]%
        {wifs/0022KC19}
\bibfield{author}{\bibinfo{person}{Xu Zhang}, \bibinfo{person}{Svebor Karaman},
  {and} \bibinfo{person}{Shih{-}Fu Chang}.} \bibinfo{year}{2019}\natexlab{}.
\newblock \showarticletitle{Detecting and Simulating Artifacts in {GAN} Fake
  Images}. In \bibinfo{booktitle}{\emph{{IEEE} International Workshop on
  Information Forensics and Security}}. \bibinfo{pages}{1--6}.
\newblock


\bibitem[Zheng et~al\mbox{.}(2022)]%
        {Zheng_2022_CVPR}
\bibfield{author}{\bibinfo{person}{Jinkai Zheng}, \bibinfo{person}{Xinchen
  Liu}, \bibinfo{person}{Wu Liu}, \bibinfo{person}{Lingxiao He},
  \bibinfo{person}{Chenggang Yan}, {and} \bibinfo{person}{Tao Mei}.}
  \bibinfo{year}{2022}\natexlab{}.
\newblock \showarticletitle{Gait Recognition in the Wild With Dense 3D
  Representations and a Benchmark}. In \bibinfo{booktitle}{\emph{{IEEE}
  Conference on Computer Vision and Pattern Recognition}}.
  \bibinfo{pages}{20228--20237}.
\newblock


\end{thebibliography}

\clearpage
\appendix
\section{Overview of the appendix}
In the appendix, we provide additional materials
% to supplement the paper
, including:
\begin{itemize}

    \item Additional samples:

        \quad Additional comparisons of heatmaps of $\overline{\rm TFC}$ like the ones shown in Fig.7.

        \quad Additional comparisons of synthetic frames like the ones shown in Fig.~5 and Fig.~6.

    \item Discussion:
    
        \quad Broader impacts of the work.
    
        \quad Limitations of the work.

\end{itemize}

\section{Additional Samples}
In this section, we provide additional samples to support the paper.

\subsection{Heatmaps}
\label{heatmaps}
\begin{figure}[!htbp]
    \centering
    \includegraphics[width=0.9\columnwidth]{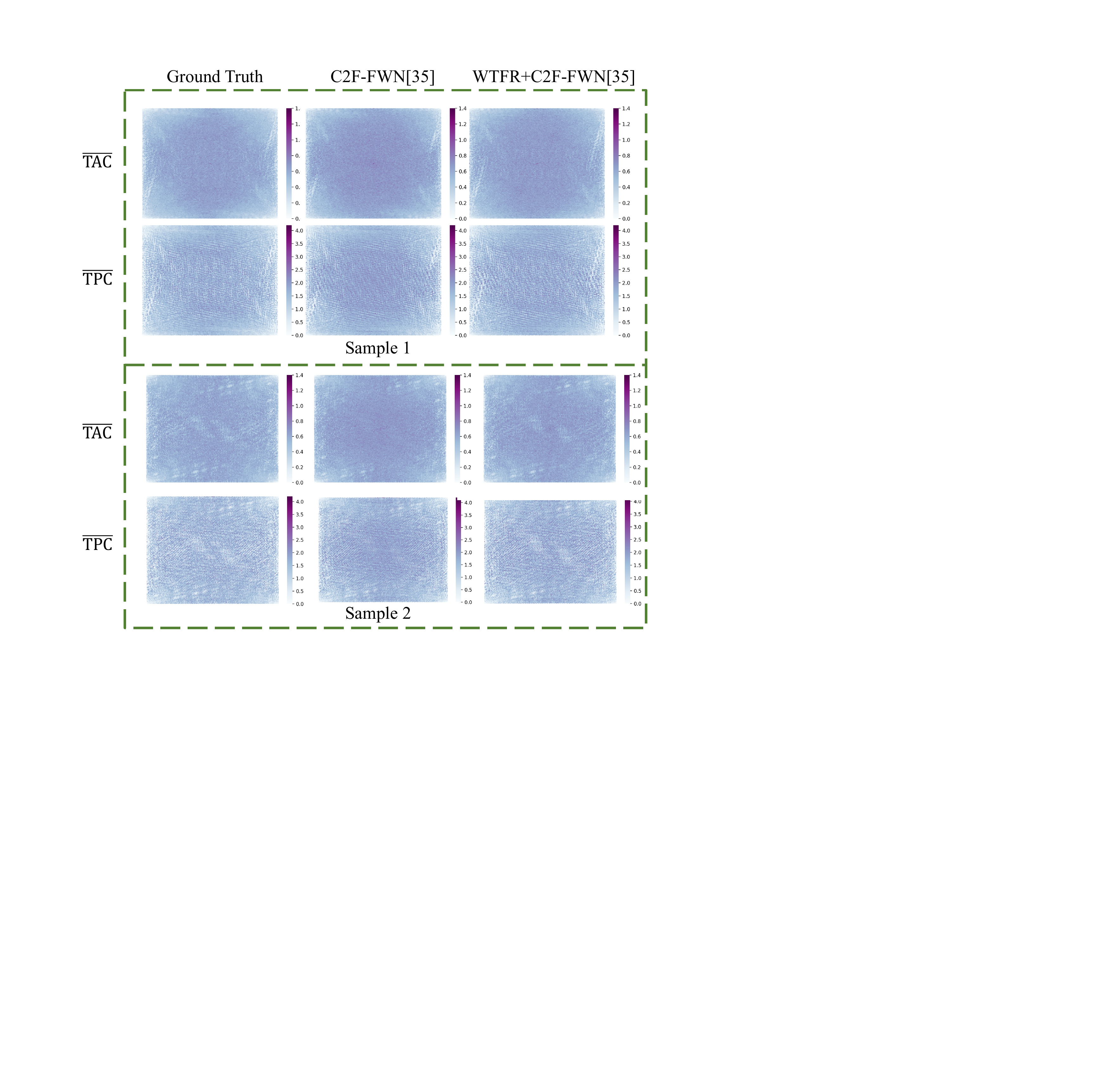}
    \caption{Additional heatmaps of mean Temporal Frequency Change $\overline{\mathbf{TFC}}$~(i.e., $\overline{\mathbf{TAC}}$ and $\overline{\mathbf{TPC}}$) from ground truth videos, synthetic videos of original models~(i.e., C2F-FWN), and synthetic videos of WTFR-enhanced models. With the proposed WTFR, the $\overline{\mathbf{TAC}}$ and $\overline{\mathbf{TPC}}$ of synthetic videos are closer to thoes of natural videos.}
    \label{additional_heatmaps}
\end{figure}

The FreMORT proposed in this paper significantly improves the temporal consistency, which 
% , which is achieved by guiding the human motion transfer methods to model the temporal frequency changes of natural videos with the proposed WTFR loss.
has been quantitatively evaluated in the Experiments section. To demonstrate that the proposed WTFR indeed narrows the discovered temporal frequency gap, we present additional examples in SUPP-Fig.~\ref{additional_heatmaps} to supplement Fig.~7.

The examples in SUPP-Fig.~\ref{additional_heatmaps} reflect that existing GAN-based methods cannot effectively model the frequency-domain distribution of natural videos with only pixel-domain supervision. In contrast, the proposed WTFR complements the gap between synthetic and natural videos in the frequency domain, which makes the $\overline{\rm TFC}$ of synthetic videos closer to that of natural videos. 

\subsection{Comparison of Synthetic Frames}

\begin{figure*}[!th]
    \centering
    \includegraphics[width=1.0\textwidth]{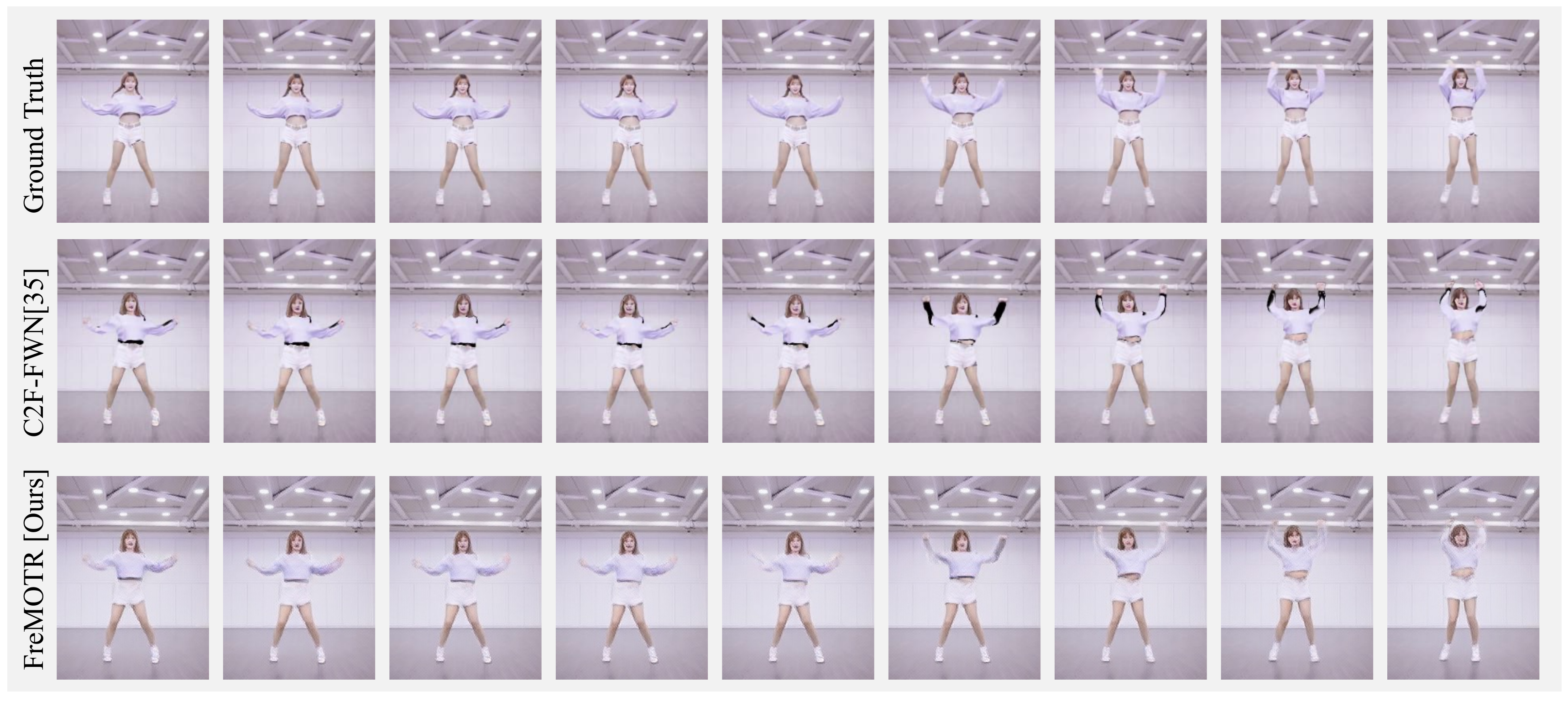}
    \caption{Frame-level comparison of video synthesized by the original model and video synthesized by the proposed FreMOTR. Both the temporal consistency and the frame-level image quality get improved with our FreMOTR.}
    \label{supp_case1}
\end{figure*}

\begin{figure*}[!th]
    \centering
    \includegraphics[width=1.0\textwidth]{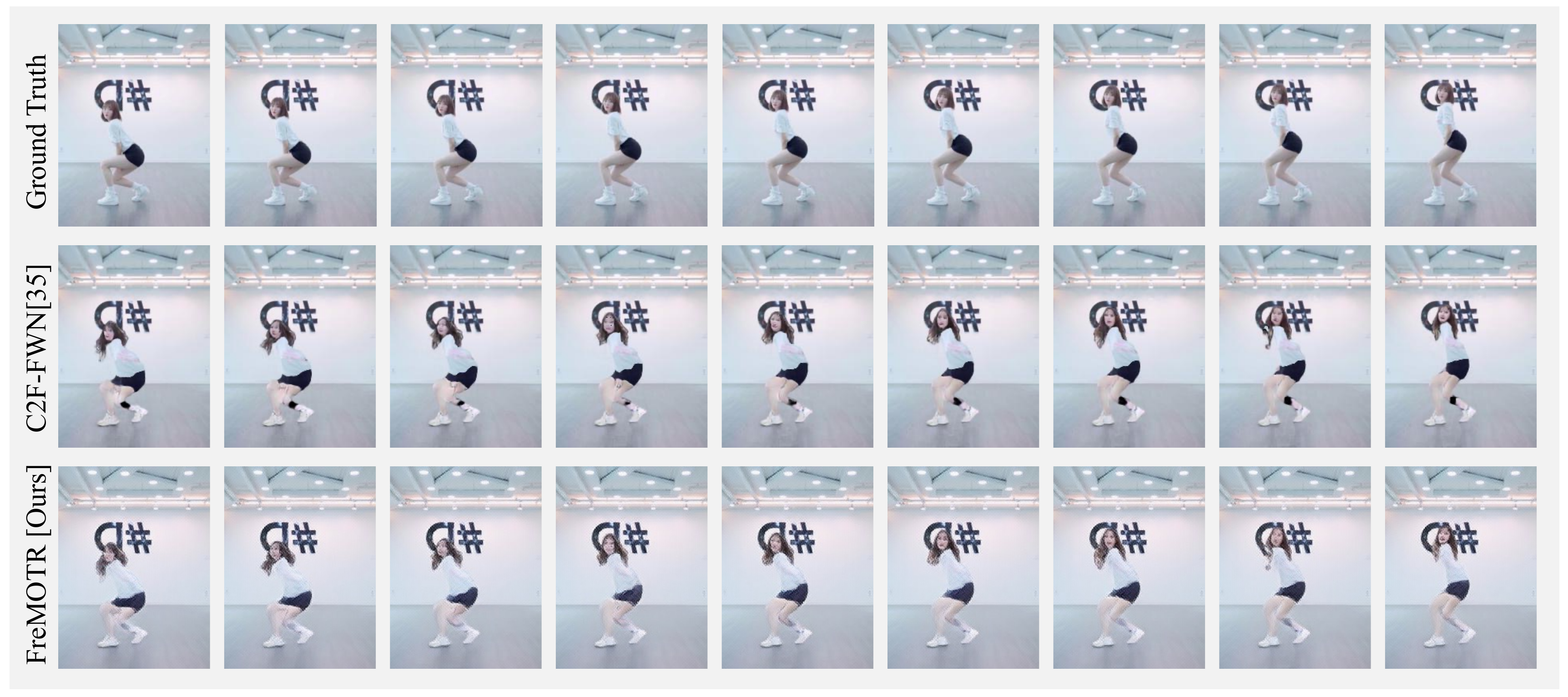}
    \caption{Additional frame-level comparison of video synthesized by the original model and the proposed FreMOTR.}
    \label{supp_case2}
\end{figure*}

Additional comparisons of synthetic frames are shown in SUPP-Fig.~\ref{supp_case1}, SUPP-Fig.~\ref{supp_case2}, and SUPP-Fig.~\ref{supp_case3}. The frames synthesized by the original model contain inconsistent components, and both the temporal consistency and frame-level image quality get improved with the proposed FreMOTR, which is identical to the quantitative analyses in the main paper. More importantly, the proposed FreMOTR can achieve a long-time consistency for successive frames.

% \begin{figure*}[!th]
%     \centering
%     \includegraphics[width=1.0\textwidth]{Figures/supp_case.pdf}
%     \caption{Frame-level comparison of video synthesized by the original model and video synthesized by the proposed FreMOTR. Both the temporal consistency and the frame-level image quality get improved with our FreMOTR.}
%     \label{supp_case}
% \end{figure*}

% \begin{figure*}[!th]
%     \centering
%     \includegraphics[width=0.8\textwidth]{Figures/supp_case_3.pdf}
%     \caption{Additional frame-level comparison of video synthesized by the original model and the proposed FreMOTR.}
%     \label{supp_case3}
% \end{figure*}

\section{Discussion}
\label{diccussion}

\subsection{Broader Impacts}
\label{broader_impact}
We propose a novel perspective to analyze and improve human motion transfer in the frequency domain. 

To narrow the temporal frequency gap, a model-agnostic method~(i.e., WTFR) is designed, which can be employed by any other GAN-based human motion transfer methods.
We also calculated the WTFR loss when reproducing the backbone without backward propagation. 
We observe that the WTFR loss decreased when the models converged gradually, and the trend of WTFR loss is similar to other losses like reconstruction loss, perceptual loss, and attention regularization loss.
This phenomenon demonstrates that the proposed WTFR is related to not only temporal consistency but also image quality. 
% The finding suggests that the proposed WTFR may also help improve image generation by removing the temporal part.

The proposed FreMOTR makes the synthetic videos more temporally consistent and photo-realistic, while the generation of videos without permissions may lead to negative social impact. 
But at the same time, the discovered gap in the frequency domain between synthetic and natural videos still exists even with the proposed FreMOTR. 
Therefore, the frequency domain analysis will be an effective tool that helps detect synthetic videos like DeepFake.

\subsection{Limitations}
\label{limitations}

We regularize the appearance conditioned on the coarse frames synthesized by the backbone, and thus the final performance is subject to the backbone. The overall appearance of the frame synthesized by the backbone may collapse due to various reasons. 
Although our FreMOTR can significantly improve the temporal consistency,  some extremely bad cases cannot be completely handled,e.g., a man without a face. In contrast, a stronger backbone that can synthesize fine single frames will benefit more from our FreMOTR.

In addition, the face and facial structure become a bit blurred, as we regularize the entire image directly. An additional face reenactment method like~\cite{eccv/ThiesETTN20} many further improves the frame quality, which is not included in the framework.

The WTFR considers temporal consistency that enhances the training process based on adjacent frames. 
Like other sequence-to-sequence tasks that consider successive inputs and outputs, the accumulated error from previous frames may influence the training process. 
The strategy to reduce the influence of error accumulation needs to be explored in the future.

\end{document}